%% file: arxiv.tex
\documentclass[10pt,twocolumn,letterpaper]{article}

\usepackage[pagenumbers]{cvpr} 


\definecolor{cvprblue}{rgb}{0.21,0.49,0.74}
\usepackage[pagebackref,breaklinks,colorlinks,allcolors=cvprblue]{hyperref}

\usepackage{makecell}
\usepackage{amsmath}
\usepackage{multirow}
\usepackage{colortbl}
\usepackage{caption}[font={small}]
\usepackage{tcolorbox}
\usepackage{xcolor}[HTML]

\def\methodname{SOLAMI}
\def\datasetname{SynMSI}

\def\paperID{3522} 
\def\confName{CVPR}
\def\confYear{2025}

\definecolor{bluecolor}{HTML}{4A86E8}
\definecolor{greencolor}{HTML}{6AA84F}
\definecolor{orangecolor}{HTML}{FF9900}

\title{
    {
    \textcolor{orangecolor}{SO}\textcolor{greencolor}{LA}\textcolor{bluecolor}{MI}: 
    \textcolor{orangecolor}{So}cial Vision-\textcolor{greencolor}{L}anguage-\textcolor{greencolor}{A}ction 
    \textcolor{bluecolor}{M}odeling \\ for \textcolor{bluecolor}{I}mmersive Interaction with 3D Autonomous Characters
    }
}

\author{
Jianping Jiang$^{* \dagger 1}$, Weiye Xiao$^{* \ddagger 1}$, \\
Zhengyu Lin$^{1}$, Huaizhong Zhang$^{1}$, Tianxiang Ren$^{1}$, 
Yang Gao$^{1}$, Zhiqian Lin$^{1}$,  \\
Zhongang Cai$^{\S 1,2}$, Lei Yang$^{\S 1}$, Ziwei Liu$^{\S 2}$ \\
$^{1}$SenseTime Research, $^{2}$S-Lab, Nanyang Technological University \\
{\small $^{*}$ Equal Contribution, $^{\dagger}$ Project Lead, $^{\ddagger}$ Engineering Lead, $^{\S}$ Corresponding Author} \\
\url{https://solami-ai.github.io/}
}

\begin{document}

\input{figures/teaser}

\input{sec/0_abstract_arxiv}    
\input{sec/1_intro}
\input{sec/2_related_work}
\input{sec/3_method}

\input{sec/4_data_collection}
\input{sec/5_VR_demo}
\input{sec/6_experiments}

\input{sec/7_conclusion}
{
    \newpage
    \small
    \bibliographystyle{ieeenat_fullname}
    \bibliography{main}
}
\clearpage
\maketitlesupplementary
\appendix

\input{sec/supp_E_limitations}
\input{sec/supp_A_method}
\input{sec/supp_B_data}

\input{sec/supp_D_experiments}
\input{sec/acknowledgements}

\end{document}

%% file: figures/teaser.tex
\twocolumn[{
    \renewcommand\twocolumn[1][]{#1}
    \maketitle
    \vspace{-34pt}
    \begin{center}
        \includegraphics[width=1\textwidth]{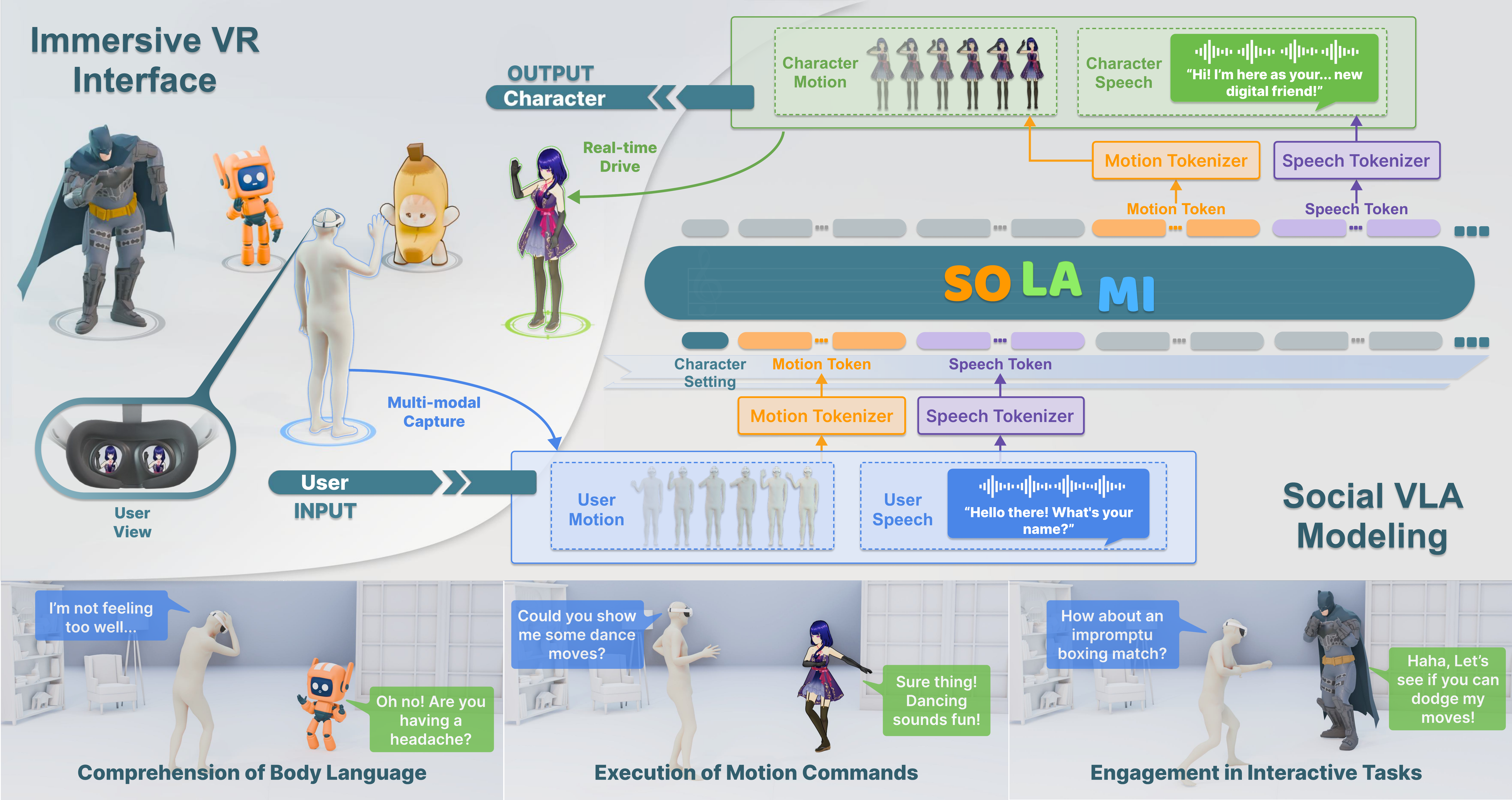}
        \captionof{figure}{
        \textbf{\methodname{}} enables the user to interact with 3D autonomous characters through speech and body language in an \textbf{immersive VR environment} via an end-to-end social vision-language-action model, which is trained on our synthesized multimodal dataset \textbf{\datasetname{}}.
        }
        \label{fig:teaser}
    \end{center}
}]

%% file: sec/0_abstract_arxiv.tex
\begin{abstract}
Human beings are social animals. How to equip 3D autonomous characters with similar social intelligence that can perceive, understand and interact with humans remains an open yet foundamental problem. 
In this paper, we introduce \textbf{\methodname{}}, the first end-to-end \textbf{So}cial vision-\textbf{L}anguage-\textbf{A}ction (VLA) \textbf{M}odeling framework for \textbf{I}mmersive interaction with 3D autonomous characters. 
Specifically, \methodname{} builds 3D autonomous characters from three aspects: \textbf{1) Social VLA Architecture:} We propose a unified social VLA framework to generate multimodal response (speech and motion) based on the user's multimodal input to drive the character for social interaction.
\textbf{2) Interactive Multimodal Data:} We present \textbf{\datasetname{}}, a \textbf{syn}thetic  \textbf{m}ultimodal \textbf{s}ocial \textbf{i}nteraction dataset generated by an automatic pipeline using only existing motion datasets to address the issue of data scarcity.
\textbf{3) Immersive VR Interface:} We develop a VR interface that enables users to immersively interact with these characters driven by various architectures. 
Extensive quantitative experiments and user study demonstrate that our framework leads to more precise and natural character responses (in both speech and motion) that align with user expectations with lower latency. 
\end{abstract}

%% file: sec/1_intro.tex
\section{Introduction}
\label{sec:intro}

Have you ever imagined having an immersive face-to-face conversation with a character you deeply admire? 
Not merely through speech dialogue, but through an interaction where you can observe its subtle facial expressions, natural body language, and even fleeting emotional changes. 
Psychology research~\cite{cummings2016immersive, slater2009inducing, VRemotion07, hudson2019or} shows that in social interactions, the greater the level of immersion, the better the human experience. 
However, current character agents~\cite{characterai, talkie, CharacterLLM} are still limited to text or voice interactions. 
This limitation prompts us to build 3D autonomous characters with richer modalities.

Developing an autonomous 3D character requires effectively modeling its behavior system, which involves two major challenges:
1) The 3D character needs to accurately observe and understand the information conveyed by the user, and respond appropriately based on the context and its character setting through speech, body motion, and facial expression, \emph{etc.} This goes beyond previous singular human-related tasks, such as motion generation~\cite{motiondiffuse}, motion understanding~\cite{motionGPTnips}, and audio-to-motion~\cite{listendenoise}.
2) Data for multimodal interactions between users and 3D characters is extremely scarce due to the prohibitive cost of the comprehensive setup.

Previous work~\cite{dlp} is primarily based on the LLM-Agent framework, using text to link various sub-modules (such as motion captioning and text-to-motion). 
While this approach performs well in high-level tasks like planning and memory, it tends to fall short in tasks such as understanding user behaviors and providing timely body motion responses. 
This limitation arises because using text as the intermediary between modules conveys high-level information but often omits subtle nuances. And the sub-modules (motion captioning, speech recognition, \emph{etc.}) in the complex engineering framework incur substantial latency that undermines the timeliness of a natural communication~\cite{cognitivearch}.

Interestingly, research on robotics shows the similar conclusion. The LLM-Agent  framework can handle planning tasks~\cite{SayCan}, but for low-level manipulation tasks, end-to-end Vision-Language-Action (VLA) models built upon LLMs show superior performance~\cite{RT-1, RT-2, LEO,openvla}. 
We argue that digital avatars are essentially robots with virtual humanoid embodiment. Therefore, building a VLA model for social interactions with users is a promising direction.

In this paper, we implement an end-to-end social VLA model, \textbf{\textit{\methodname{}}}.
Our model, built upon a decoder-only LLM backbone, processes the inputs of user speech and motion into discrete representations, and generate the responsive speech and motion tokens, which are then decoded to the character's speech and motion.
This modeling approach can effectively learn character behavior patterns across motion and speech modalities and offer low latency.

Although there are numerous datasets related to human social behaviors~\cite{Audio2Photoreal, dlp, inter-x, intergen}, comprehensive multimodal interaction datasets remain scarce.
Thus, we introduce a data synthesis method that utilizes existing text-motion datasets to automatically construct multimodal interaction data at a low cost.
Leveraging our extensively curated topics (5.3 K), uniformly processed motion database (46 K), and iterative script refinement pipeline, we develop \textbf{\textit{\datasetname{}}}, a dataset containing 6.3 K multi-turn multimodal conversation items.
To evaluate the effectiveness of our method, we developed a VR interface where users can immersively interact with various 3D characters. 
Quantitative experimental results and user study analysis show that our approach produces more precise and natural social interaction experience with lower latency.

To summarize, we contribute \textbf{1) A new VLA architecture} to model the character's behavior system for immersive social interaction; \textbf{2) A dedicated synthesizing pipeline} that automatically generates large-scale multimodal interactive dataset, \datasetname{}; \textbf{3) An immerse VR interface} for users to interact with various characters through speech and motion.  

%% file: sec/2_related_work.tex
\section{Related Work}
\label{sec:related_work}

\subsection{Human Motion \& LLM}

Existing role-play agents primarily rely on text~\cite{CharacterLLM}, speech~\cite{characterai, talkie, gpt-4o}, or video~\cite{bodyofher} as interactive media, but building 3D autonomous characters means modeling 3D embodied behavior, especially body language. 
Unlike single-purpose human motion tasks~\cite{motiondiffuse, listendenoise, inter-x, intergen, smpler-x, EgoLM, duolando}, we expect this 3D character not only to comprehend user speech and body language but also to respond according to its profile setting. 
Large language models (LLMs)~\cite{llama2, gpt-4o}, with their remarkable emergent abilities~\cite{emergent}, provide promising solutions for this direction.
One approach~\cite{motiongpt-jiang, motiongpt-zhang, avatargpt, chatpose} tries to integrate LLMs and human motion in an end-to-end fashion, enabling a single model to perform multiple motion-related tasks. However, the goal of these methods is not to generate responsive motion based on the input motion according to the character setting. 
Another approach~\cite{dlp, chathuman} utilizes the LLM as a versatile brain center that controls various sub-modules (e.g., motion understanding, text-to-motion generation) with text or code instructions. 
However, this modular approach inherently introduces information loss and engineering-related time latency. 
Therefore, how to create feasible autonomous 3D characters for immersive interaction remains an open challenge.
  
\subsection{Embodied Intelligence}
Building 3D autonomous characters means creating virtual humanoid agents with embodied intelligence. 
In the field of embodied intelligence, researchers have found the high-level abilities (planning, memory, etc.) of LLM-Agents across various environments, including factories~\cite{SayCan, palm-e}, 2D sandbox settings~\cite{generative-agents}, and 3D gaming spaces~\cite{voyager}. 
For tasks requiring low-level skills, such as manipulation, end-to-end VLA models~\cite{LEO, RT-1, RT-2, openvla} have shown considerable potential. 
Despite existing efforts~\cite{ego4d, egoexo4d, EgoLM} in egocentric human-related tasks, the capabilities of VLA models in social interactive tasks with humans have not been fully explored.

\input{figures/pipeline}

%% file: figures/pipeline.tex
\begin{figure*}[t]
    \centering
    \includegraphics[width=\linewidth]{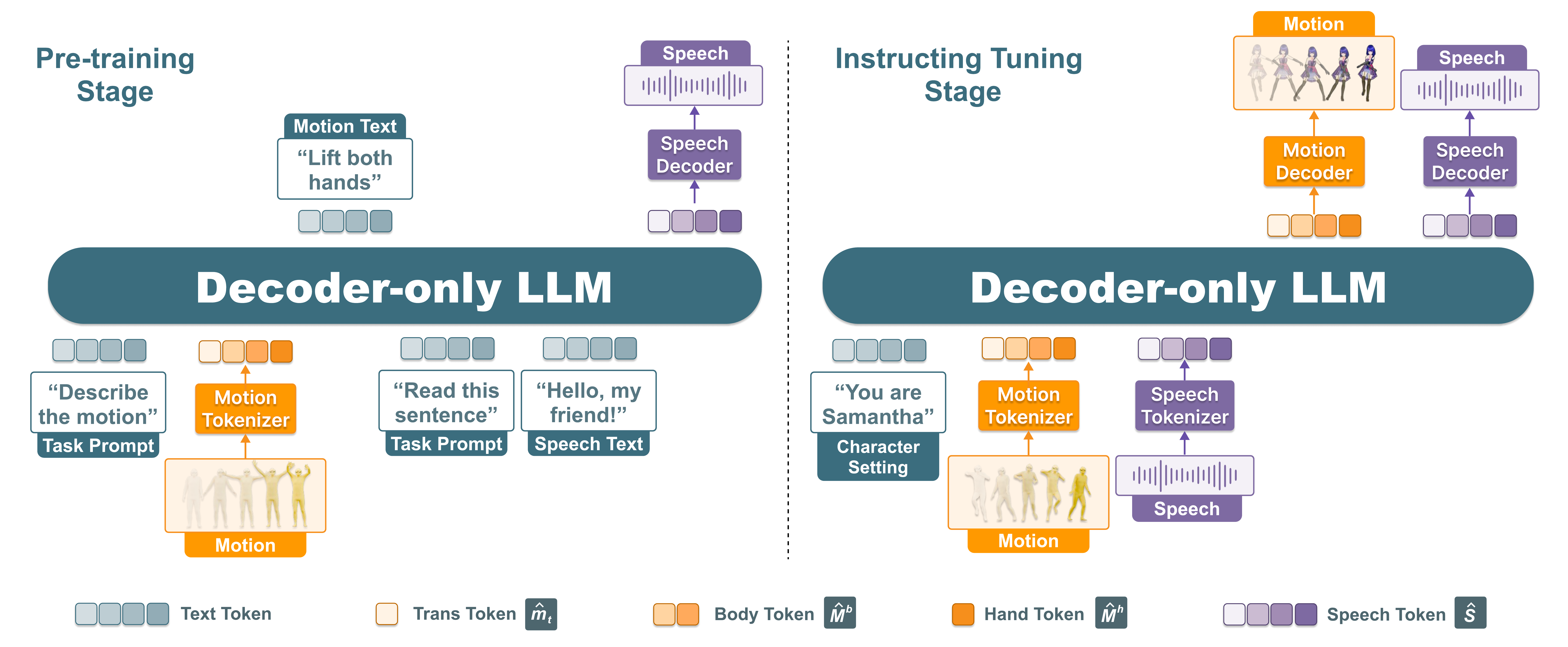}
    \caption{
    Training pipeline of \methodname{}. We train \methodname{} through a three-stage process. In the pre-training stage, we train the model with motion-text and speech-text related tasks to align the speech and motion modalities with language. During the instruction tuning stage, we train the model with social multimodal multi-round interaction data, enabling it to generate multimodal responses that align with the character settings and the context of the topic.
    }
    \label{fig:pipeline}
\end{figure*}

%% file: sec/3_method.tex
\section{Social Vision-Language-Action Modeling}
\label{sec:method}

\subsection{Architecture}

Our framework is an end-to-end social VLA model that takes the user's speech and motion as input and generates the character's responsive speech and motion as output. 
In this process, speech and motion are added as new languages to the LLM text vocabulary.
First, the user's speech and motion are converted into discrete motion tokens and speech tokens via a motion tokenizer and a speech tokenizer, respectively. 
A decoder-only LLM backbone then predicts the character's output motion and speech tokens based on the user's input tokens and the character setting. 
The generated tokens are subsequently decoded into corresponding motion and speech by their respective decoders.

\noindent \textbf{Motion Representation.}
To take advantage of SMPL-X's~\cite{smplx} compatibility with industry animation workflow, we directly model human poses as SMPL-X joint rotations instead of keypoint positions~\cite{motiongpt-jiang, humantomato} to facilitate animation of characters down the stream.

\noindent \textbf{Motion Tokenizer.}
Our motion tokenizer employs a Vector Quantized Variational Autoencoders (VQ-VAE) structure as \cite{motiongpt-jiang, motiongpt-zhang}. It learns discrete representations of motion, enabling the LLM to understand the text-motion connection. 
We design separate VQVAEs ($Q^{b}$, $Q^{h}$, $Q^{t}$) for the body motion $\textbf{m}^{b}$, hand motion $\textbf{m}^{h}$, and inter-character relative transform (rotation and translation) $\textbf{m}_{t}$ for higher reconstruction accuracy~\cite{humantomato}. 
The quantization process for motion $\textbf{m}^{u}$ is formulated as
\begin{equation}
    {\hat{m}}^{u}_t = Q^{u}(\textbf{m}^{u}_{t}) = \arg \min_{z_i \in \mathbb{Z}_{u}}{\lVert \textbf{m}^{u}_{t} - z_i \rVert_{2}},
\end{equation}
where $\mathbb{Z}_u$  is the codebook of motion part $u \in \{b, h, t\}$, and ${\hat{m}}^{u}_t$ is the corresponding motion tokens.
The VQVAEs of body and hand, $Q^{b}$ and $Q^{h}$, apply 1D convolutions on motion features along the temporal dimension to get $L_M$ sequential tokens $\hat{M}^b = [{\hat{m}}^{b}_1, {\hat{m}}^{b}_2, ..., {\hat{m}}^{b}_{L_M}]$ and $\hat{M}^h = [{\hat{m}}^{h}_1, {\hat{m}}^{h}_2, ..., {\hat{m}}^{h}_{L_M}]$, and the VQVAE $Q^{t}$ uses MLPs to get the transform token ${\hat{m}}^{t}$ of a sequence. 

\noindent \textbf{Speech Tokenizer.}
Research~\cite{audiolm, speechgpt} on speech discretization mainly utilizes the RVQ-VAE structure~\cite{soundstream}. 
In this work, we utilize the SpeechTokenizer~\cite{speechtokenizer} that disentangles semantic and acoustic information within speech $S$. 
This allows us to use the semantic tokens $\hat{S} = [{\hat{s}}^{s}_1, {\hat{s}}^{s}_2, ..., {\hat{s}}^{s}_{L_S}]$ from the first layer as input to the LLM, reducing inference costs ($L_S$ is the sequence length of semantic tokens). 
Simultaneously, a short sample of the character’s voice (4 to 6 seconds) can be used as a prompt to achieve instance voice cloning when decoding through SoundStorm~\cite{soundstorm}.

\noindent \textbf{Multi-modal Multi-round Interaction.}
User-character interaction is formulated as a multi-round conversation fashion of common LLMs~\cite{gpt-4o, llama2}.
When the user sends speech and motion to \methodname{}, the model auto-regressively generates speech and motion responses based on previous dialogue contents and the character setting.
To facilitate training, we use special tokens to mark the start and end of each modality sequence as \cite{videopoet, motiongpt-jiang}.
The interaction process can be formulated as follow:

\begin{tcolorbox}[title={Interaction Template}]
\label{tb: interaction template}
Input: \\
System Prompt: \textless Character\_Placeholder\textgreater \\
User:  \textless M\_Placeholder\textgreater \textless S\_Placeholder\textgreater \\
Character: \textless M\_Placeholder\textgreater \textless S\_Placeholder\textgreater \\
... \\
User: \textless M\_Placeholder\textgreater \textless S\_Placeholder\textgreater \\
Output:\\
Character: \textless M\_Placeholder\textgreater \textless S\_Placeholder\textgreater
\end{tcolorbox}
\noindent
where \textless Character\_Placeholder\textgreater \  is placeholder for character's text description, \textless M\_Placeholder\textgreater \ for motion token sequences, and \textless S\_Placeholder\textgreater \ for speech token sequences.



\subsection{Training}
The training of \methodname{} adopts a three-stage strategy. 

\noindent \textbf{Stage 1: Tokenizer Training.}
The training approach for the motion tokenizer uses the fashion of \cite{motiongpt-jiang}. The train loss is 
\begin{equation}
   \mathcal{L}_m =  \lambda_r \mathcal{L}_r + \lambda_e \mathcal{L}_e + \lambda_c \mathcal{L}_c + \lambda_v \mathcal{L}_v,
\end{equation}
where $\mathcal{L}_r$ means reconstruction loss, $\mathcal{L}_e$ embedding loss, $\mathcal{L}_c$ commitment loss, $\mathcal{L}_v$ velocity loss, and $\lambda_{*}$ are manually adjusted weights.
For the speech tokenizer, we use the pre-trained checkpoint from AnyGPT~\cite{anygpt}.
We freeze the tokenizers' weights after this stage.

\noindent \textbf{Stage 2: Multi-task Pre-training for Modality Alignment.}
As shown in \cref{fig:pipeline}, the second stage is multi-task pre-training, achieving modality alignment between motion and text, as well as between speech and text.
It is necessary because motion data is scarce, and direct training on multimodal interaction data results in suboptimal models, as demonstrated in subsequent ablation studies. 
For motion and text alignment, we use 46 K motion-text pairs for text-to-motion generation and motion captioning tasks, and 11 K interactive motion pairs for two-person motion generation.
To align the speech and text, we train the model with 410 K speech-text pairs for text-to-speech and automatic speech recognition tasks, and 100 K speech dialogue pairs for speech-to-speech generation.
The tasks are formulated as ``\textit{User: \textless Task\_Placeholder\textgreater \textless Input\_Modality\_Placeholder\textgreater;\quad Character:\textless Output\_Modality\_Placeholder\textgreater} ".
To balance the scale disparity between the motion and speech data, we sampled them at a 4:6 ratio during training.

\noindent \textbf{Stage 3: Instruction Tuning for Multi-turn Conversation.}
In the third stage, we perform instruction tuning by applying supervised fine-tuning with multimodal interaction data, enabling the model to handle long-sequence, multi-turn dialogues, as shown in \cref{fig:pipeline}.
We utilize 5.7 K multimodal conversation items for supervised fine-tuning.
Each conversation item is fed to the model in the format as the Interaction Template in \cref{tb: interaction template}.
We experiment with two approaches: full-parameter fine-tuning and LoRA fine-tuning~\cite{lora}.
We supervised only the character's response to teach the model how to react to the user's behavior.
Thus we train the model by maximizing the log-likelihood of the next-token prediction and the train loss is:
\begin{equation}
\begin{aligned}
    \mathcal{L}_{\text{IT}} = & - \sum_{r=1}^{R} \sum_{i=1}^{L_{M}^r} \text{log} p_{\Theta}(\hat{m}^{r}_i | \hat{m}^{r}_{i-1}, ..., \hat{m}^{r}_{1}, \hat{S}_{\textless r}, \hat{M}_{\textless r}) \\ & - \sum_{r=1}^{R} \sum_{i=1}^{L_{S}^r} \text{log} p_{\Theta}(\hat{s}^{r}_i | \hat{s}^{r}_{i-1}, ..., \hat{s}^{r}_{1}, \hat{S}_{\textless r}, \hat{M}_{\leq r} ),
\end{aligned}
\end{equation}
where $\Theta$ is the network parameter of the LLM backbone or LoRA weights, $R$ is the conversation round, $\hat{S}_{r}$ and $\hat{M}_{r}$ are the $r$-th round speech and motion token sequences with lengths $L_{M}^r$ and $L_{S}^r$, respectively.

%% file: sec/4_data_collection.tex
\section{\datasetname{} Dataset}
\label{sec:data_collection}

\input{figures/data_synthesis}

Social interaction between users and virtual characters is inherently unique, which makes collecting such multimodal interaction data particularly challenging. 
Currently, available public datasets~\cite{humanml3d, inter-x, Audio2Photoreal} are incomplete for our task.
To address this issue, we propose a data synthesis pipeline that leverages existing motion-text datasets, text-based role-play models, and speech synthesis methods and generates a large-scale multimodal dialogue dataset, \textbf{\datasetname{}}.

\subsection{Motion Data}
We collect motion-text data for two purposes: first, to achieve alignment between motion and text during pre-training, and second, to generate multimodal data for instruction tuning. 
Since our work focuses on modeling social interactions, we select existing datasets that contain rich social actions: HumanML3D~\cite{humanml3d} with 24 K motion-text pairs, Inter-X~\cite{inter-x} with 20 K motion-text pairs and 10 K two-person motion pairs, and DLP-MoCap~\cite{dlp} with 2 K motion-text pairs.
Since the Inter-X~\cite{inter-x} dataset contains only text descriptions of two-person interactive motion without descriptions of individual motion, we used GPT-4o~\cite{gpt-4o} to decompose the two-person action descriptions into single-person motion-text pairs.
Additionally, we used GPT-4o~\cite{gpt-4o} to synthesize comprehensive text descriptions for each motion clip by consolidating multiple possible descriptions, thereby providing more detailed textual annotations that preserve motion details.

\subsection{Speech Data}
We use speech-text data for speech-text alignment in the pre-training stage.
Speech datasets involve CommonVoice~\cite{commonvoice} (150 K speech-text pairs in our experiments), AnyInstruct~\cite{anygpt} (200 K speech-text pairs and 100 K speech-to-speech items), and synthetic speech data (60 K speech-text pairs) by text-to-speech approaches (Azure TTS and XTTS\_v2~\cite{xtts}).

\subsection{Multimodal Data Synthesizing}
Previous data synthesis methods~\cite{videochat, llava, minigpt-4} usually use the general abilities of advanced LLMs~\cite{gpt-4o} and text-annotated multimodal data to generate synthetic data. 
However, generating social multimodal interaction data has not yet been achieved. 
This challenging task requires high-quality expression of body language, voice consistency that matches the characters, and suitable dialogue content.

As shown in \cref{fig:data synthesis}, our synthesizing pipeline includes 4 steps.
(1) First, we collect 5.3 K character-related and daily topics from internet platforms (Google Trends~\cite{googletrends}, Zhihu~\cite{zhihu}, Jike) and brainstorms of GPT-4o~\cite{gpt-4o} to improve the diversity of the dialogue contents.
(2) Based on the topic, character setting, and previous round of scripts, we use GPT-4o~\cite{gpt-4o} to generate textual descriptions (motion, speech, expression \emph{etc.}) for the next round of the dialogue. 
(3) Then we utilize the text embedding~\cite{textembedding} of the motion description to retrieve the most relevant motions from our meticulously curated motion-text database. 
(4) Moreover, we refine the generated speech text with the retrieved motions to ensure that speech and motion are well-coordinated. 
The motion database with detailed text annotations and the refinement process can alleviate the misalignment between the real motion and speech in the LLM-Agent method~\cite{dlp}.
By iteratively repeating this process, we can generate multi-round dialogue contents across many characters, where the motions are sourced from the motion database, and the speeches are synthesized using TTS/voice cloning (Azure TTS and XTTS\_v2~\cite{xtts}) to maintain consistency with the character’s voice style.
We finally obtained 6.3 K multimodal dialogue items.

%% file: figures/data_synthesis.tex
\begin{figure*}[t]
    \centering
    \includegraphics[width=\linewidth]{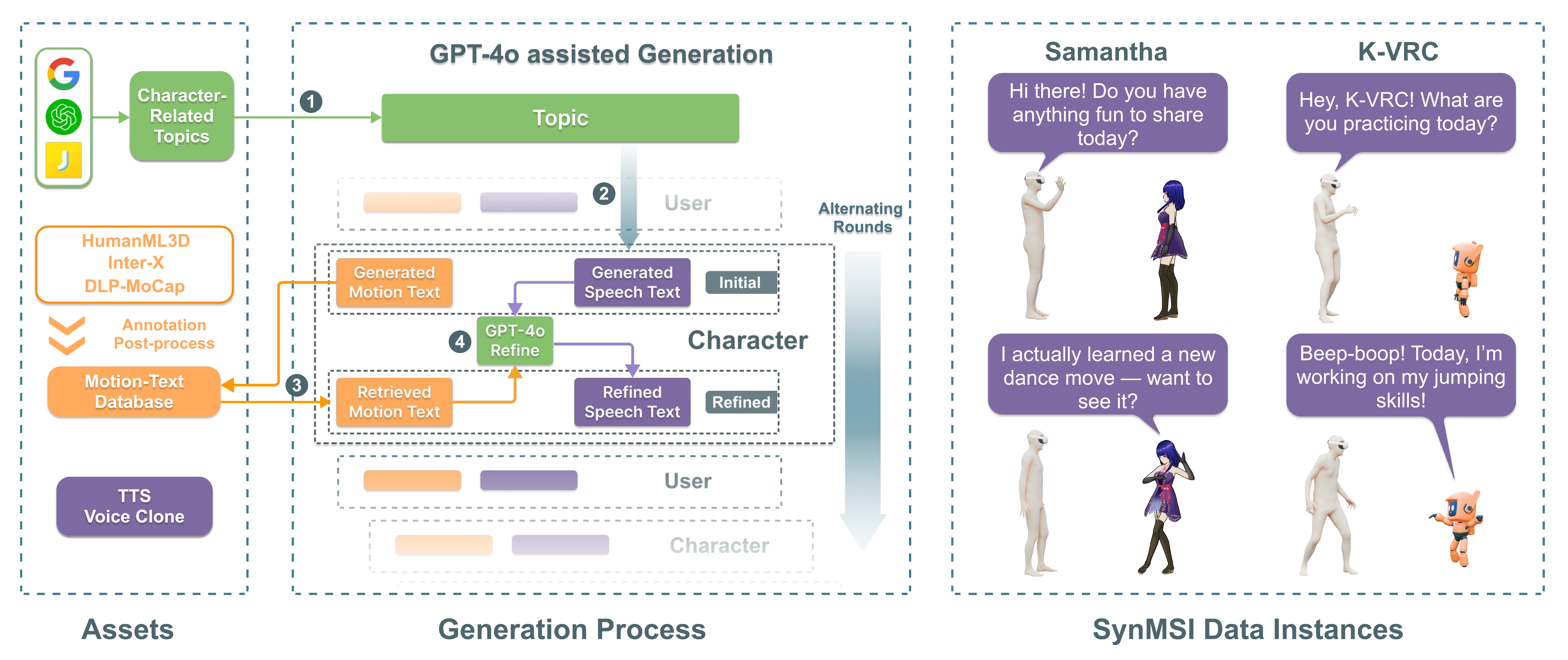}
    \caption{
    \datasetname{} dataset generation.
    Our synthesizing pipeline consists of 4 steps.
Based on numerous character-relevant topics and state-of-the-art LLMs~\cite{gpt-4o}, we generate text scripts for multimodal dialogues. Using a large-scale motion database~\cite{inter-x, humanml3d, dlp}, we retrieve the most appropriate motions and refine the speech scripts accordingly. Finally, we employ TTS/voice cloning~\cite{xtts} to generate character-specific speech. This approach enables us to create multimodal interaction data of various characters using only existing motion datasets.
    }
    \label{fig:data synthesis}
\end{figure*}

%% file: sec/5_VR_demo.tex
\section{VR Interface}
\label{sec:vr_demo}

To demonstrate our method directly, we developed a VR interface with an Oculus Quest 3 frontend and a backend service, as shown in \cref{fig:vr demo}. 
The frontend enables immersive interaction between users and 3D autonomous characters, while the backend, powered by 2 H800 GPUs, supports the computation of various baselines in our experiments. 
During usage, the VR headset captures the user’s speech and body motion, and sends them to the backend computation nodes. 
For motion capture, we use the Quest’s full-body tracking system~\cite{questsim} to obtain pose parameters, which are then retargeted onto an SMPL-X model~\cite{smplx}. 
The computation nodes generate the body motion parameters and speech responses of the character based on the multimodal input. 
Then we apply an audio-to-face method, UniTalker~\cite{unitalker}, to generate the character’s facial animation parameters. 
The facial and body parameters are jointly retargeted onto a 3D character model~\cite{synbody}, completing one cycle of social interaction.
For a natural user experience, we employ preset idle motions on the character when the method is LLM+Speech or the character is waiting for the user's input.

\noindent \textbf{3D Character Assets.}
Our 3D character portfolio covers a diverse range of entities, including AI assistant avatars, famous cinematic roles, internet memes, and real-world celebrity personas. These 3D models are sourced from open-source repositories under \textit{CC Attribution-NonCommercial-ShareAlike License} as well as our manual creation using VRoid Studio~\cite{vroid}.
We subsequently employ facial rigging, skinning, bone chain simulation, retargeting, and texture and material creation in Unity Engine processes to yield functional 3D character assets.

\input{figures/vr_demo}

%% file: figures/vr_demo.tex
\begin{figure}[t]
    \centering
    \includegraphics[width=\linewidth]{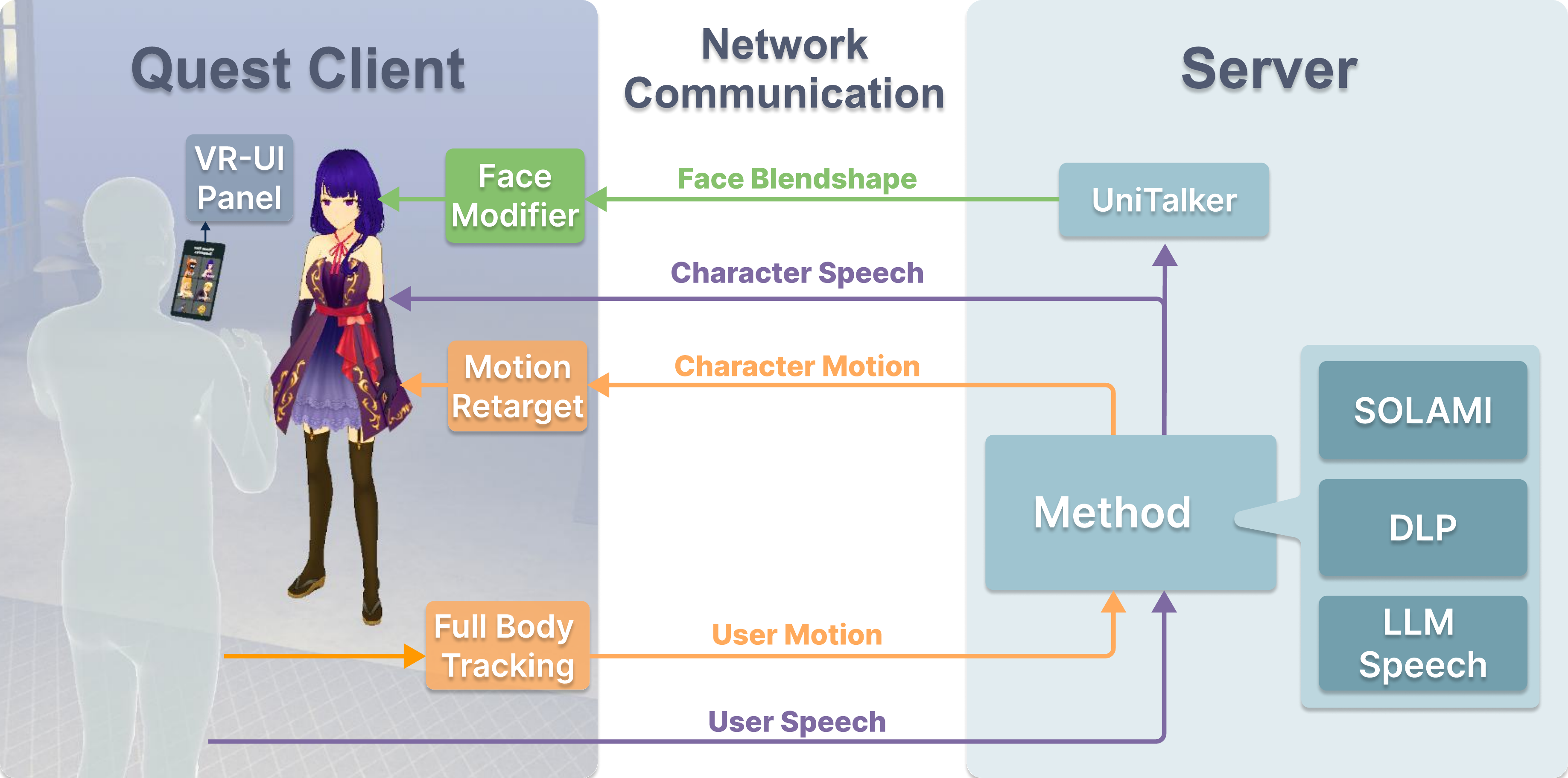}
    \caption{
    VR interface architecture. Our VR project consists of a Quest 3 client and a server. The Quest client captures and transmits user body motion and speech to the server. The server then generates character's speech, body motion, and face blendshape parameters based on the selected methods. The response is then sent back to the Quest client to drive the character.
    }
    \label{fig:vr demo}
\end{figure}

%% file: sec/6_experiments.tex
\section{Experiments}
\label{sec:experiments}


\input{tables/quant_method}

\subsection{Experimental Settings}
\label{sec: experimental settings}

In our experiment, we selected the AnyGPT-base model~\cite{anygpt} (based on LLaMA2-7B~\cite{llama2}) as the backbone for \methodname{}, because it is an open-source model available at our experimental time that supports end-to-end speech processing. 
During the pre-training stage, we utilize 32 V100 GPUs to train the model for 3 K steps (batch size 256, learning rate 4e-5). 
For instruction tuning, we train the \methodname{} for 800 steps using 16 V100 GPUs (batch size 48, learning rate 2e-5).
For LoRA fine-tuning~\cite{lora}, we set the rank as 8 and alpha as 16.
We split the synthesized multimodal data into training and test sets with a 9:1 ratio. 
We use DeepSpeed~\cite{deepspeed} to accelerate the training.
During testing, we evaluate each round of the character’s response.

\noindent \textbf{Baselines.}
To validate the performance improvement in social interaction brought by incorporating 3D modalities (such as body motion), we compared \methodname{} with the \textbf{\textit{LLM+Speech}} and the \textbf{\textit{AnyGPT (fine-tune)}} approach. 
For the \textit{LLM+Speech} framework, the user's speech is first transcribed into text using ASR techniques, which is then fed to a LLM to generate the character’s response in text, and subsequently converted into speech using TTS. 
For fairness, we use LLaMA2-7B-Chat~\cite{llama2} as the LLM backbone, Whisper large-v3~\cite{whisper} for ASR, and XTTS\_v2~\cite{xtts} for voice cloning.
For the \textit{AnyGPT (fine-tune)} framework, we use the speech data of \datasetname{} to train the AnyGPT-base model~\cite{anygpt} with the same parameter settings.
To compare the effectiveness of the LLM-Agent architecture with the social VLA framework, we used \textbf{\textit{DLP}}~\cite{dlp} as a baseline method. 
In \textit{DLP} framework, the user's speech and body motion are separately processed as text descriptions by ASR and motion captioning modules.
Based on the input text descriptions, LLM generates the character's text instructions of speech and motion, which are transferred into speech and body motion by TTS and motion generation module.
The speech component of \textit{DLP} follows the \textit{LLM+Speech} method. 
Considering that MoMat-MoGen module in \textit{DLP} is too slow for user interaction (over 5 seconds latency), we use MotionGPT~\cite{motiongpt-jiang} for motion captioning and motion generation. 
To ensure a fair comparison, we used the same motion data as the pre-training stage of \methodname{} to train MotionGPT.
Additionally, we conducted ablation experiments on the effect of the pre-training stage, marked as \textbf{\textit{(w/o pretrain)}}. 
For the ablation study of the motion tokenizer, please refer to the supplementary materials.
We use vLLM~\cite{vllm} to accelerate the LLM backbones for low latency interaction.

\subsection{Quantitative Evaluation}
\label{sec: quatitative evaluation}

We conducted quantitative evaluation for our method and all the baselines mentioned in \cref{sec: experimental settings}.

\noindent \textbf{Evaluation Metrics.}
For motion, we evaluate the model responses using metrics including FID, diversity, PAMPJPE (mm), and angle error~\cite{EgoLM}. 
Following Duolando~\cite{duolando}, we obtain FID and diversity using motion features from AIST++~\cite{aist++}.
For speech, we use VC similarity~\cite{uniaudio} to evaluate the voice similarity with the character.
To evaluate the content quality of speech, we first use Whisper-large-v3~\cite{whisper} to transcribe the speech into text. 
Then following \cite{CharacterLLM, charactereval}, we employ GPT-4o~\cite{gpt-4o} as the judge to assess \textbf{\textit{Context Relevance}} and \textbf{\textit{Character Consistency}} on a Likert scale ranging from 1 to 5. 
\textit{Context Relevance} indicates whether the speech content aligns with the topic and context of the conversation, while \textit{Character Consistency} assesses whether the content adheres to the character settings.
For inference latency (seconds), we deploy all the models on 2 H800 GPUs with vLLM~\cite{vllm} framework and asynchronous mechanisms to improve performance while maintaining  fairness.

\noindent \textbf{Quantitative Results.}
The quantitative results in \cref{tab:quant_method} demonstrate that, using the same foundation model (Llama2)~\cite{llama2} as the backbone, our method, \methodname{} (full params), significantly outperforms other methods in terms of motion quality and inference latency.

\noindent \textbf{Motion Quality}. \methodname{} demonstrates superior performance compared with the DLP method~\cite{dlp} across multiple motion metrics.
This is because the social VLA model achieves comprehensive modality alignment among speech, motion, and language through training on our high-fidelity character-specific multimodal data.
Our model can precisely perceive the user's physical motions and linguistic cues, enabling semantically rich interactive motions in response.
This ability contrasts with the LLM-Agent architectures of the DLP method, which exhibits limitations in conveying multimodal nuances through text-only intermediary representation.

\noindent \textbf{Speech Quality}. Our method demonstrates the capability to synthesize a voice tone that matches the character with a higher Voice Cloning (VC) Similarity score.
Our model also shows better performance on the context relevance of the dialogue than other methods.
Because the inclusion of the motion modality enables the model to perceive the user's body language, while the LLM+Speech or AnyGPT(fine-tune) method lacks this capability.
In terms of character consistency, our model achieves secondary performance metrics.
We suspect this may be due to the incorporation of motion and speech modalities, which potentially affects the character-related knowledge embedded within the original LLM.
The performance degradation is similar to the observations in ~\cite{vila, palm-e}.

\noindent \textbf{Inference Latency}. 
Our end-to-end approach is significantly superior to the modular pipeline approaches (LLM+Speech or DLP).
Because the end-to-end VLA model naturally aligns with the process of real-time human communication.
Theoretically, if we could collect data on real-time interactions between humans and characters, our method could achieve full-duplex streaming interaction.

\noindent \textbf{Ablation Study.}
As shown in \cref{tab:quant_method}, the pre-training stage of \methodname{} leads to better performance in both motion and speech quality.
We believe that the pre-training stage, which aligns motion, speech, and language, facilitates the model's ability to learn the multimodal dialogue skill during the instruction tuning stage.
Instruction tuning using LoRA~\cite{lora} shows weaker results compared to full parameter fine-tuning.
We think that the gap between the data distribution of pre-training tasks and the multimodal instruction tuning task is substantial, and LoRA fine-tuning alone is insufficient for the model to learn the strong multimodal dialog ability.

\input{tables/questionaire}

\input{figures/user_study}

\input{figures/qualitative}

\subsection{VR Interface User Study}
\label{sec: user study}
Quantitative evaluation of a single modality alone cannot fully compare the 3D autonomous characters built on different frameworks. 
To address this, we conducted user study with a VR interface. 
As shown in \cref{fig:vr demo}, we employ LLM+Speech, DLP (MotionGPT), and \methodname{} as method options of the server backend and the same VR frontend across different methods.
This implies that the variable in the experiment is the driving method in the server.
Users are asked to engage in more than five rounds of dialogue with the VR character before completing our questionnaire.

\noindent \textbf{Evaluation Metrics.}
The indicators and corresponding questions for our questionnaire are shown in  \cref{tab:questionnaire}. 
\textit{Motion Coherence} evaluates whether the character’s motion aligns with the character setting and dialogue content. 
\textit{Motion Interaction} assesses whether the character can understand the semantics of the body motion and effectively interact with the user. 
\textit{Speech Coherence} examines whether the generated speech aligns with the context and character settings. 
And the \textit{Overall Experience} measures the user’s satisfaction with the overall experience.
Each question is rated on a 1 to 5 Likert scale, with higher scores indicating greater satisfaction. 
The score for each dimension is calculated as the average score of its corresponding questions.
We ultimately collected 60 survey responses, with participants from various gender and age groups.

\noindent \textbf{Results.}
As shown in the \cref{fig:user study}, our method achieved the highest user satisfaction across all dimensions. 
\methodname{} demonstrates superior performance over the DLP method across all metrics, validating the effectiveness of an end-to-end social VLA model in character behavior modeling. 
While the DLP method shows lower speech consistency compared to the LLM+Speech method, it excels in motion-related metrics and achieves higher overall satisfaction, indicating that effective body language can enhance user experience despite speech quality limitations.

To provide a more intuitive demonstration of our model's capabilities, we replay the user study experiments and render them, as shown in the \cref{fig:qualitative}.
The results indicate that \methodname{} demonstrates excellent capabilities in body language understanding, motion command execution, and body interaction.
For better understanding, we also present the workflow from a first-person view during actual usage. We encourage readers to explore our supplementary materials and video for a more detailed overview of the model experiments, data generation pipeline, VR interface construction process, and comprehensive experimental results.

%% file: tables/quant_method.tex
\begin{table*}[ht]
\centering
\caption{\textbf{Quantitative results of baselines and \methodname{}}. `$\uparrow$'(`$\downarrow$') indicates that the values are better if the metrics are larger (smaller). We run all the evaluations 5 times and report the average metric. The best results are in bold and the second best results are underlined.} 
\label{tab:quant_method}
\setlength{\tabcolsep}{2mm}
\resizebox{\textwidth}{!}{
\small
\begin{tabular}{lcccccccc}
\hline

\multirow{2}{2cm}{\centering Methods} & \multicolumn{4}{c}{\centering Motion Metrics} & \multicolumn{3}{c}{\centering Speech Metrics} & \multirow{2}{2cm}{\centering \thead{Inference\\ Latency}$\downarrow$}\\
& FID$\downarrow$ & Diversity$\uparrow$ & PA-MPJPE$\downarrow$ & Angle Error$\downarrow$ & VC Similarity$\uparrow$ & Context Relevance$\uparrow$ & Character Consistency$\uparrow$ \\
\hline
\datasetname{} Dataset & - & 9.136 & - & -  & - & 4.888 & 4.893 & - \\
\hline

LLM+Speech (Llama2)~\cite{llama2} & - & - & - & - & 0.818 & 3.527 & \textbf{3.859} & 3.157 \\

AnyGPT (fine-tune)~\cite{anygpt} & - & - & - & - & 0.819 & 3.502 & 3.803 & \textbf{2.588} \\


DLP (MotionGPT)~\cite{dlp} & \underline{4.254} & 8.259 & 165.053 & 0.495 & 0.812 & \underline{3.577} & 3.785 & 5.518 \\

SOLAMI (w/o pretrain) & 5.052 & \underline{8.558} & \underline{159.709} & \underline{0.387} & \underline{0.820} & 3.541 & 3.461 & 2.657 \\

SOLAMI (LoRA)  & 15.729 & 8.145 & 167.149 & 0.400 & 0.770 & 3.251 & 3.423 & 2.710 \\

SOLAMI (full params)  & \textbf{3.443} & \textbf{8.853} & \textbf{151.500} & \textbf{0.360} & \textbf{0.824} & \textbf{3.634} & \underline{3.824} & \underline{2.639} \\

\hline
\end{tabular}}
\end{table*}

%% file: tables/questionaire.tex
\begin{table}[t]
\centering
\small
\caption{Questionnaire settings of our user study.}
\label{tab:questionnaire}
\resizebox{\linewidth}{!}{
\begin{tabular}{cl}
\hline
Dimension & Questions \\
\hline
\multirow{2}{*}{Motion Coherence} & Does the motion match the character's setting? \\
 & Does the motion align well with speech? \\
\hline
\multirow{2}{*}{Motion Interaction} & Does the character follow motion instructions correctly? \\
 & Does the character understand user's motion? \\
 \hline
 \multirow{2}{*}{Speech Consistency} & Does the speech match the character's setting? \\
 & Is the speech relevant to the current topic? \\
  \hline
  Overall Experience & How would you rate the overall experience? \\
\hline
\end{tabular}}
\end{table}

%% file: figures/user_study.tex
\begin{figure}[t]
    \centering
    \includegraphics[width=\linewidth]{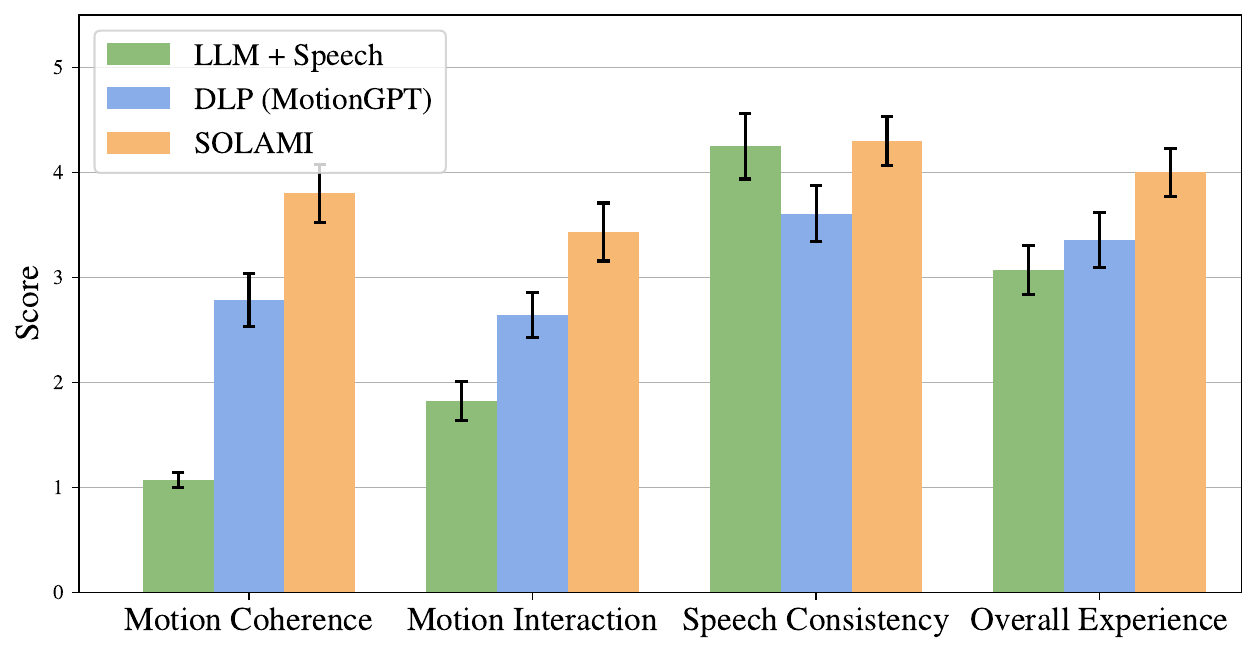}
    \caption{
    Results of the user study with 95\% confidence.
    }
    \vspace{-10pt}
    \label{fig:user study}
\end{figure}

%% file: figures/qualitative.tex
\begin{figure*}[t]
    \centering
    \includegraphics[width=\linewidth]{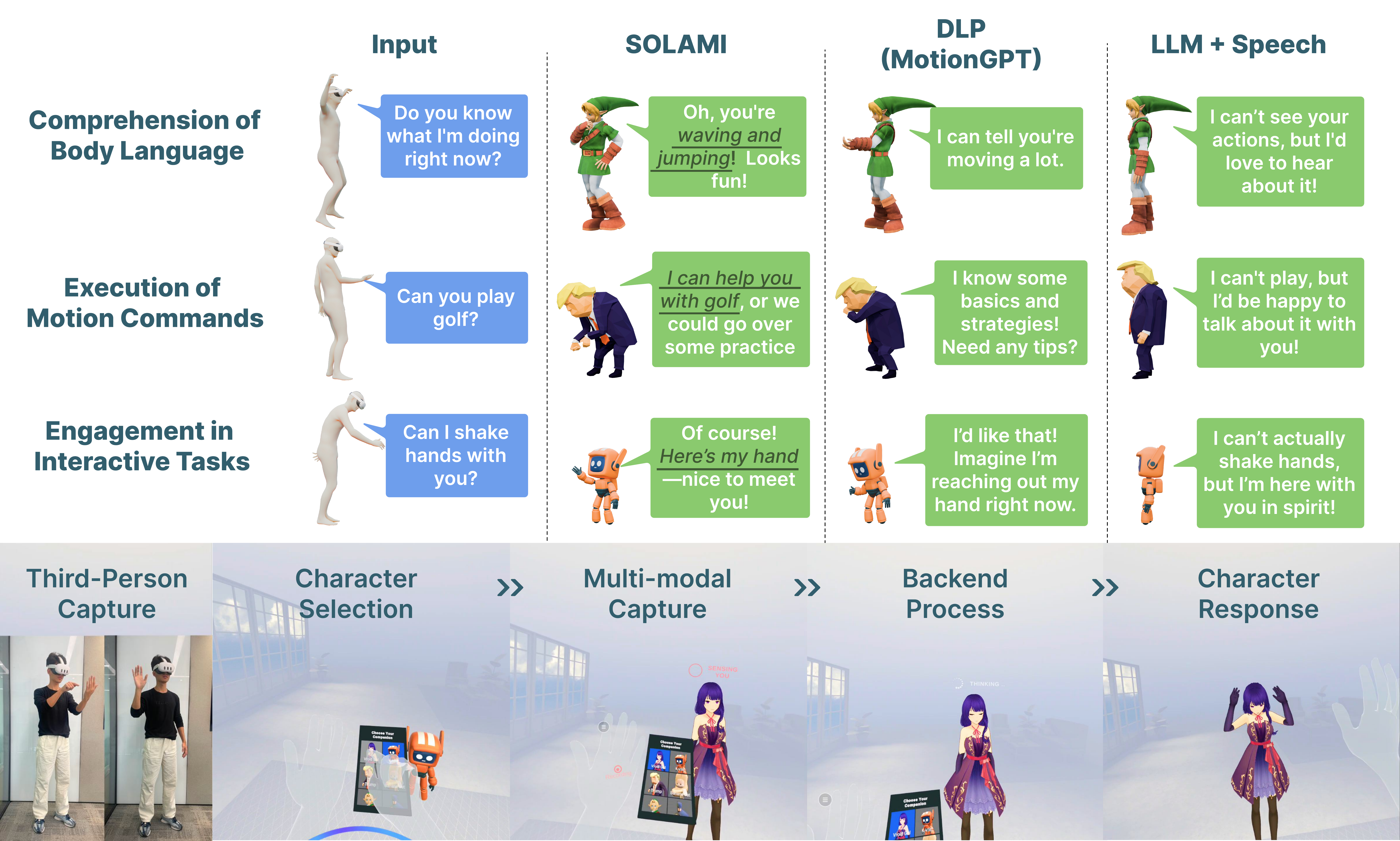}
    \caption{
    Qualitative results of \methodname{} and baselines, and the user workflow for VR experience.
    Our social VLA model, trained in an end-to-end strategy on \datasetname{} dataset, can accurately perceive the semantic information embedded within users' speech and motion input, and subsequently generate natural and coherent responses.
    }
    \vspace{-15pt}
    \label{fig:qualitative}
\end{figure*}

%% file: sec/7_conclusion.tex
\vspace{-2pt}
\section{Conclusion}
\label{sec:conclusion}

In this paper, we propose \methodname{}, an approach for building 3D autonomous characters. 
This approach includes three key components: \textit{1) Architecture:} A novel social VLA modeling framework enabling multimodal social interaction; \textit{2) Multimodal Data Synthesizing:} A pipeline for automatically generating multimodal interaction data from existing incomplete datasets; \textit{3) VR Interface:} A VR engineering framework that facilitates immersive interactions between users and various characters. 
Together, these modules contribute to an enhanced user interaction experience.

%% file: sec/supp_E_limitations.tex
\section{Future Work}
\label{sec: supp: limitations}

Our work, \methodname{}, represents a preliminary exploration for building 3D autonomous characters. 
While it has performed well in comparative experiments, there remains significant room for improvement on aspects as follows:

\begin{itemize}
    \item \textbf{Input Modality:} For dyadic social interaction, using the user's body motion and speech as input is sufficient. However, when considering multi-person interaction or interaction involving the environment and objects, video~\cite{palm-e, RT-2} or dynamic 3D scenes~\cite{shapellm} might be a better choice;
    \item \textbf{Data Collection:} Our synthetic dataset, \datasetname{}, enables satisfactory user evaluation results. However, collecting real-time data of actual dyadic interaction could enable our model to generate more precise and natural body language and speech, while also supporting duplex streaming conversations, similar to \cite{bodyofher,glm4voice}. Compared to text and video modalities, the collection of embodied 3D data is undoubtedly challenging. Potential solutions include: capturing~\cite{smpler-x} or learning human behavioral data~\cite{gen2act} from existing video datasets, building immersive interaction platforms~\cite{vrchat} to gather data on human interactions, and using surrogate control to collect data from human interactions with 3D characters~\cite{cheng2024tv};
    \item \textbf{Cross Embodiment:} Using a unified SMPL-X~\cite{smplx} model to represent characters' motion inevitably introduces challenges in cross-embodiment for different characters. While some degree of error and misalignment may not hinder information exchange in social language interaction, such representations clearly lack generalizability for fine-grained tasks (\emph{e.g.}, handshaking, object manipulation). The challenges of retargeting in 3D human-related tasks and cross-embodiment in robotics~\cite{RT-2} share similarities, providing opportunities for mutual inspiration and methodological exchange;
    \item \textbf{Long-Short Term Design:} Although \methodname{} demonstrates effective modeling for real-time interactions, its architecture encounters challenges such as computational redundancy, forgetting, and training difficulties during extended social interactions. A promising direction~\cite{dlp, fastslow} to explore is integrating long-term memory, knowledge, and skills with short-term real-time interaction. This approach could ensure interaction quality while reducing computational overhead and simplifying the training process;
    \item \textbf{Efficient Learning Method:} Although our dataset, \datasetname{}, tries to collect large-scale motion data, the inherently long-tail distribution~\cite{longtailed} of human motions results in some behaviors having very low occurrence frequencies~\cite{inter-x, humanml3d, human3.6M}. In particular, the data volume for signature actions of 3D characters is inherently limited. While models like GPT-3~\cite{gpt-3} have demonstrated remarkable few-shot learning capabilities, the data-intensive training required is currently unsustainable in the field of digital humans. Therefore, exploring effective learning methods is essential. Leveraging character-focused knowledge embedded in existing foundation models~\cite{languagereward, gpt4v-evaluation} or incorporating human evaluators~\cite{intructGPT} to guide the model in learning new skills from a small number of samples are promising research directions.
\end{itemize}


%% file: sec/supp_A_method.tex
\section{More Details of Architecture Design}
\label{sec: supp: architecture design}

\input{tables/pretrain_evaluation}

\input{tables/motion_tokenizer}

In this section, we discuss the input and output modalities of \methodname{} in \cref{sec: supp: input and output modalities}, compare the motion representation in \cref{sec: supp: motion representation}, and introduce details of our motion tokenizer and pre-training design in \cref{sec: supp: motion tokenizer and pre-training}.

\subsection{Input and Output Modalities}
\label{sec: supp: input and output modalities}

Our ultimate goal is to establish a unified behavioral modeling system for any character, where input modalities include a wide range of sensory observations, including vision, audio, and haptics \emph{etc.}, and output modalities represent actions in the finest possible granularity. 
However, currently, we need to balance the ideal with the constraints of existing data and devices to develop a model that provides an optimal user experience.

Regarding devices, we employ VR headsets instead of mobile phones or computers because VR headset enables a more immersive interactive experience by capturing and presenting richer information.

In terms of input modalities, while 3D scenes or videos could serve as input and have some foundational models~\cite{shapellm, videochat}, collecting corresponding social interaction data is challenging. 
For instance, datasets like Ego4D~\cite{ego4d} and Ego-Exo4D~\cite{egoexo4d} capture first-person videos and motion data but include very limited social interaction content and no data involving character interaction. 
Within VR environments, the majority of incremental information a character can observe comes from user's behaviors that VR devices can capture. 
Consequently, we chose user motion and speech as the primary input for \methodname{}.

Similarly, for easy synthetic data generation and model training, we maintain the same types of output modalities for the character as for the user's input. 
This symmetry ensures alignment between what the model observes and what it produces, facilitating a more natural and precise interactive experience.

\subsection{Motion Representation Comparison}
\label{sec: supp: motion representation}

Common representations of human motion are often based on 3D keypoints~\cite{humanml3d, motiongpt-jiang, humantomato}, which provide higher precision compared to methods based on joint rotations. 
However, this approach is inconsistent with the driving mechanism of 3D engines such as Unity Engine. 
When the model generates 3D keypoints, retargeting is necessary to derive the relative rotation of each joint with respect to its parent joint. 
Considering human motion priors, a typical approach~\cite{smplify} involves fitting an SMPL-X~\cite{smplx} model to the 3D keypoints using optimization strategies, and subsequently retargeting the fitted SMPL-X model to the character.
However, this process has two main drawbacks:
\begin{enumerate}
    \item \textbf{Time-Consuming Fitting Process:} The fitting step is computationally intensive. With optimized methods like SMPLify~\cite{smplify}, achieving an adequate result requires about 1 second of iteration on a V100 GPU.
    \item \textbf{Fitting Artifacts and Distortion:} Inevitable fitting errors can lead to biologically implausible joint rotations, significantly degrading visual quality.
\end{enumerate}
In our experiments, we observed that while human motion representation based on 3D keypoints performs well in terms of motion metrics, as shown in \cref{tab: supp: pre-train} and \cref{tab: supp: tokenizer}, its visual fidelity is inferior to representation based on joint rotations. 
To address this, we adopted a cont6d representation for joint rotations, achieving improved visual outcomes.

\subsection{Motion Tokenizer and Pre-training}
\label{sec: supp: motion tokenizer and pre-training}

After processing as described in \cref{sec: supp: motion representation}, we obtained a 315-dimensional motion representation. When converting this motion representation into tokens via the tokenizers, several issues need to be discussed. Should body and hand motion features be represented separately? If so, how should their tokens be handled? Should the tokens for the body and hand motions be interleaved, or should they be input as independent sequences in the pre-training stage?

Considering our computational cost, we conducted ablation experiments on the text-to-motion task using the GPT-2~\cite{gpt-2} backbone as the baseline model. 
Finally, we compared the models under the same settings using Llama2-7B~\cite{llama2} as the backbone.

As shown in \cref{tab: supp: tokenizer} and \cref{tab: supp: pre-train}, compared to unified representations of hand and body motion (marked as ``bind"), the separate representation (marked as `separate") achieves better performance, particularly with higher precision on the text-to-motion task (t2m).
However, the trade-off is that the probability of GPT-2~\cite{gpt-2} producing outputs that conform to the expected format (marked as ``Pred Valid") decreases. However, this issue is mitigated in large part by using Llama2~\cite{llama2} as the backbone model.
We think this improvement is due to the differences in the language models: GPT-2, the relatively smaller language model, has weaker comprehension of textual instructions. 
In contrast, Llama2, trained on extensive corpora, demonstrates significantly stronger text understanding capabilities.
Moreover, compared to interleaved tokens (``Yes" for ``Token Interleaved"), separate sequence representations (``No" for ``Token Interleaved") achieve better motion metrics. 
We hypothesize that this is because learning separate sequences reduces the overall complexity of the motion pre-training task, thereby improving performance.

Based on the above experimental evaluations, we ultimately select Llama2-7B~\cite{llama2} for its strong text comprehension capabilities as the LLM backbone. 
For processing motion representation, we employ separate motion tokenizers that convert the motion representation into noninterleaved token sequences. This configuration is used for the final instruction fine-tuning stage.

%% file: tables/pretrain_evaluation.tex
\begin{table*}[ht]
\centering
\caption{\textbf{Quantitative results of pre-training on text-to-motion task}. `$\uparrow$'(`$\downarrow$') indicates that the values are better if the metrics are larger (smaller). The best results are in bold and the second best results are underlined.} 
\label{tab: supp: pre-train}
\setlength{\tabcolsep}{2mm}
\small
\begin{tabular}{ccccccccc}
\hline

\multirow{2}{*}{\centering ID} & \multirow{2}{*}{\centering Body \& Hand} & \multirow{2}{*}{\centering Repre} & \multirow{2}{*}{\centering Backbone} & \multirow{2}{*}{\centering \thead{Token\\ Interleaved}} & \multicolumn{4}{c}{\centering Metrics}\\
 & & & & & FID$\downarrow$ & Diversity$\uparrow$ & PA-MPJPE$\downarrow$ & Pred Valid$\uparrow$\\
\hline
1 & bind & joints & GPT-2 & -  & \textbf{1.48} & 9.03 & 148.00 & \underline{0.836} \\
2 & bind & rotation & GPT-2 & -  & 3.44 & \underline{12.94} & 143.70 & 0.813 \\
3 & separate & rotation & GPT-2 & Yes  & 3.00 & 11.64 & 117.26 & 0.676 \\
4 & separate & rotation & GPT-2 & No  & 2.72 & \textbf{14.05} & \underline{112.53} & 0.638 \\
5 & separate & rotation & Llama2 & No  & \underline{1.82} & 10.40 & \textbf{110.23} & \textbf{0.999} \\
\hline
\end{tabular}
\end{table*}

%% file: tables/motion_tokenizer.tex
\begin{table}[t]
\centering
\caption{\textbf{Quantitative results of Motion VQVAE}. `$\uparrow$'(`$\downarrow$') indicates that the values are better if the metrics are larger (smaller). The best results are in bold.} 
\label{tab: supp: tokenizer}
\small
\begin{tabular}{ccccc}
\hline

\multirow{2}{*}{\centering ID} & \multirow{2}{*}{\centering Body \& Hand} & \multirow{2}{*}{\centering Repre} & \multicolumn{2}{c}{\centering Motion Metrics} \\
 & & & PA-MPJPE$\downarrow$ & FID$\downarrow$ \\
\hline
1 & separate & joints &  87 & \textbf{1.0} \\
2 & bind & joints & \textbf{80} & 1.3 \\
3 & separate & rotation & 88 & 1.88 \\
4 & bind & rotation & 113 & 2.34 \\
\hline
\end{tabular}
\end{table}

%% file: sec/supp_B_data.tex
\input{tables/data_collection}

\section{More Details of Data Generation}
\label{sec: supp: data generation}

In this section, we first discuss several methods for collecting multimodal social interaction data in \cref{sec: supp: comparison data collection}. Then, we introduce the technical details of \datasetname{} generation pipeline in \cref{sec: supp: data generation pipeline}.

\subsection{Comparison of Data Collection Methods}
\label{sec: supp: comparison data collection}
From the perspective of data sources, we discuss three sources: internet videos, Immersive VR platform, and existing incomplete motion capture datasets, as shown in \cref{table: data collection}.

\noindent \textbf{Collecting from Internet Videos.}
The development of mobile devices has led to an explosion of video content, and researchers naturally expect the model to learn knowledge and capabilities from internet videos. Many works aim to implicitly learn human capabilities from videos~\cite{GR-1, GR-2}, but for our task, we anticipate obtaining explicit multi-modal interactive data through various tools~\cite{gpt-4o, smpler-x}. 
Human motions can be captured through video motion capture, but current video motion capture~\cite{smpler-x} faces challenges such as occlusion, temporal discontinuity, and long-tail problems, making it difficult to obtain high-quality motions. 
Understanding and annotating human behaviors in videos can be achieved using Vision-Language Models (VLM)~\cite{gpt-4o, llava}, and we find that with appropriate post-processing these annotations are usable. 
Additionally, there is another issue: the data obtained through this method lacks first-person view and is often fixed at a third-person view, which presents challenges in perspective transformation.

\noindent \textbf{Collecting from VR Platforms.}
Building a VR interaction platform to directly collect user interaction data is the most straightforward method. 
However, two key problems arise: 1) Current VR devices' body tracking systems~\cite{questsim} cannot provide ground truth-level data. For instance, existing VR devices estimate lower body postures instead of capturing with wearable sensors, and tracking becomes unreliable when hands move beyond the sensor range of VR equipment. 2) Human interaction data differs from 3D character representations. Specifically, animated characters' movements tend to be more exaggerated compared to real human motions, which naturally introduces a data distribution gap.

\noindent \textbf{Collecting from Existing Incomplete Datasets.}
Due to the novelty of our task, there is no dataset that perfectly suits our needs. 
Common open-source datasets~\cite{humanml3d, inter-x, beat} typically provide semantic annotations for motion sequences or co-speech gestures. 
The most cost-effective and convenient approach is to complete these datasets or use them to synthesize multimodal social interaction datasets. 
However, this faces several challenges: How can we ensure the diversity of dialogue content? How can we ensure that synthesized speech and motion are reasonable? Can synthetic data guarantee user satisfaction? 
We address these questions in Sec. 4 and Sec. 6.3 of the main paper. 
And we will introduce some technical details about data synthesizing latter.

In summary, obtaining data from the internet has high potential, but current video motion capture technology is insufficient to realize this potential, and it also involves perspective transformation challenges. 
Data collection from VR platforms is limited by hardware capabilities and faces difficulties in replicating character behaviors. 
Synthesizing data based on existing datasets represents an optimal choice when balancing cost and effectiveness.

\subsection{Details of \datasetname{} Generation Pipeline}
\label{sec: supp: data generation pipeline}

\noindent \textbf{Motion Post-process}
Existing motion-text datasets~\cite{inter-x, dlp} primarily provide semantic-level text annotations, often overlooking behavioral details (such as sitting versus standing positions, orientations, \emph{etc.}). 
Considering GPT-4o's capability~\cite{gpt-4o} in understanding human behaviors in videos, as shown in \cref{table: data collection}, one approach would be to render all motions into videos and then use VLM for annotation. 
However, for a small research team, the cost of VLM API calls is relatively high. 
We propose a compromise strategy: combining multiple text annotations for a single motion and using GPT-4o~\cite{gpt-4o} to generate a comprehensive, detailed description. 
In practice, we find this method to be quite effective.

\input{figures/topics}

\noindent \textbf{Topics Collection.}
Without topic guidance, conversations with LLMs often converge to simple, generic content rather than character-specific, in-depth content~\cite{dlp, CharacterLLM}. Using prompts to guide conversation is a common strategy. We collected topics from the following perspectives:
\begin{enumerate}
    \item Character-related topics: These topics are difficult to collect in bulk from the internet and were generated through GPT-4o~\cite{gpt-4o} brainstorming;
    \item News-related topics: Google Trends~\cite{googletrends} has compiled many news topics that people care about in daily life;
    \item Daily life topics: Some community websites, such as Jike, specifically curate such topic content;
    \item Topics people are curious about: Common Q\&A websites (such as Quora, Zhihu~\cite{zhihu}) specifically organize these topics.
\end{enumerate}
After collecting these topics, we used LLMs to post-process them, filtering and organizing them into topics suitable for character conversation.
Topic keywords are shown in \cref{fig: supp: topics}.

\noindent \textbf{Task Generation.}
Beyond daily conversation content, we also want \methodname{} to learn direct understanding of human body language and the ability to explicitly follow human instructions. 
For this purpose, when synthesizing data, we set up different tasks in the system prompt:
\begin{itemize}
    \item \textbf{common:} daily conversation;
    \item \textbf{motion understanding:} requires users to generate motions with strong semantic information, and the character can clearly express understanding of body movements;
    \item \textbf{instruction following:} requires users to give clear motion instructions, and the character can output corresponding instructed movements;
    \item \textbf{imitation:} requires the character to imitate user's motion.
\end{itemize}

\noindent \textbf{Script Generation Methods.}
Since we are using the chat version of LLMs, we experimented with and compared three script generation strategies:
\begin{enumerate}
    \item Method 1: Round-by-Round completion: Using LLM to complete and refine the speech and motion text for each character round by round, which is the method mentioned in our main paper.
    \item Method 2: Character Agent Dialogue: Similar to the SocioMind approach~\cite{dlp}, using two LLMs to play two roles (User and Character), and alternately outputting speech and motion text, followed by refinement.
    \item Method 3: One-shot generation: Generating the entire multi-turn dialogue script at once, then revising the script round by round based on retrieved motions.
\end{enumerate}
According to our experimental results, Method 1 and Method 2 produce better results. 
Although Method 3 initially generates good scripts, the quality deteriorate after multiple rounds of modifications during motion-text database alignment. 
To produce \datasetname{}, we randomly alternate between Methods 1 and 2 to generate text scripts.

\noindent \textbf{Interactive Motion.}
If we only use single-person motions, our model would lack the capability for two-person interaction. 
To address this issue, during script generation, when we retrieve a motion of one person in an interactive motion, we ask the LLM whether to use the motion of another person from the same interactive motion when generating the next round of motion text.

%% file: tables/data_collection.tex
\begin{table*}[t]
    \centering
    \caption{\textbf{Methods of collecting multimodal interaction data.}}
    \resizebox{\textwidth}{!}{
    \begin{tabular}{ccc}
    \toprule
   Methods & Input & Output\\
   \hline
    \multirow{2}{*}{\thead{MoCap Human \\ Motions from \\ Internet Videos\\ with SMPLer-X~\cite{smpler-x}}}
    &   \begin{minipage}[b]{0.9\columnwidth}
		\centering
        \raisebox{-.05\height}
        {\includegraphics[width=1.\linewidth]{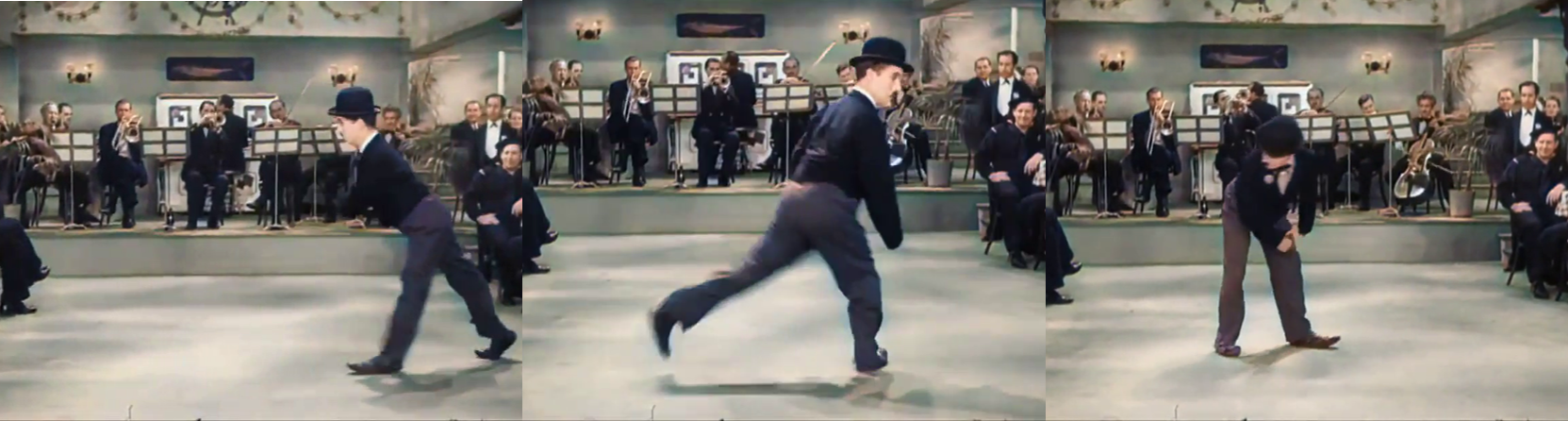}}
	\end{minipage} & \begin{minipage}[b]{0.9\columnwidth}
		\centering
        \raisebox{-.05\height}
        {\includegraphics[width=1.\linewidth]{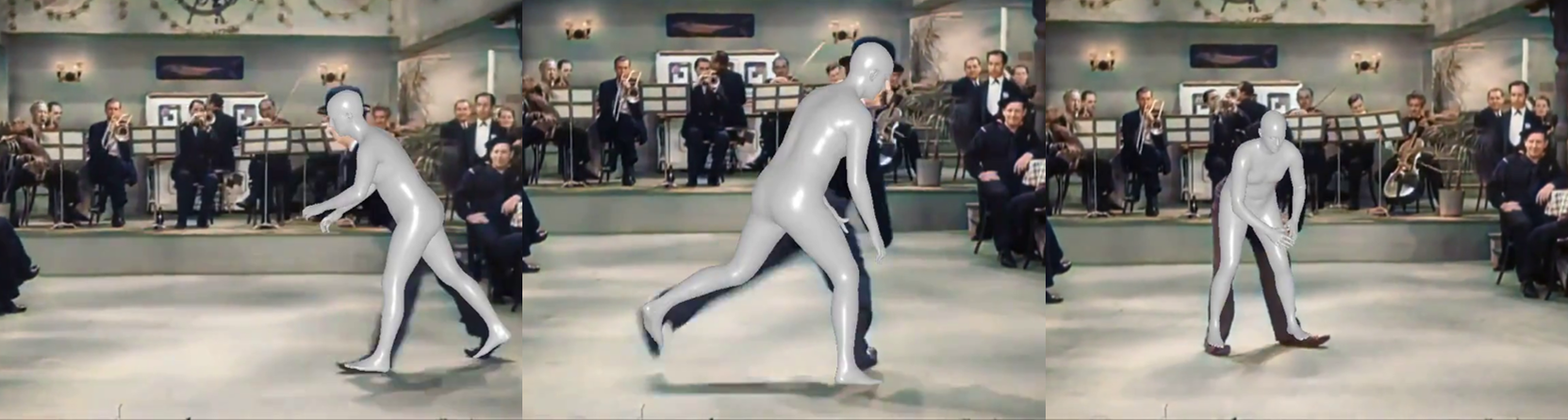}}
	\end{minipage}\\
    
    &  \begin{minipage}[b]{0.9\columnwidth}
		\centering
        \raisebox{-.05\height}
        {\includegraphics[width=1.\linewidth]{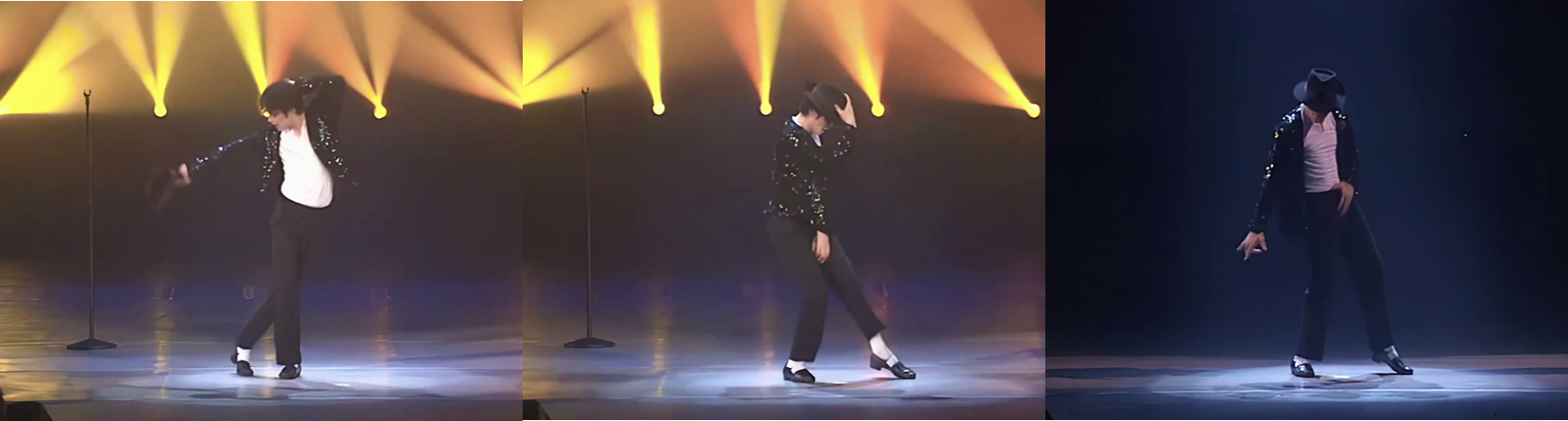}}
	\end{minipage} & \begin{minipage}[b]{0.9\columnwidth}
		\centering
        \raisebox{-.05\height}
        {\includegraphics[width=1.\linewidth]{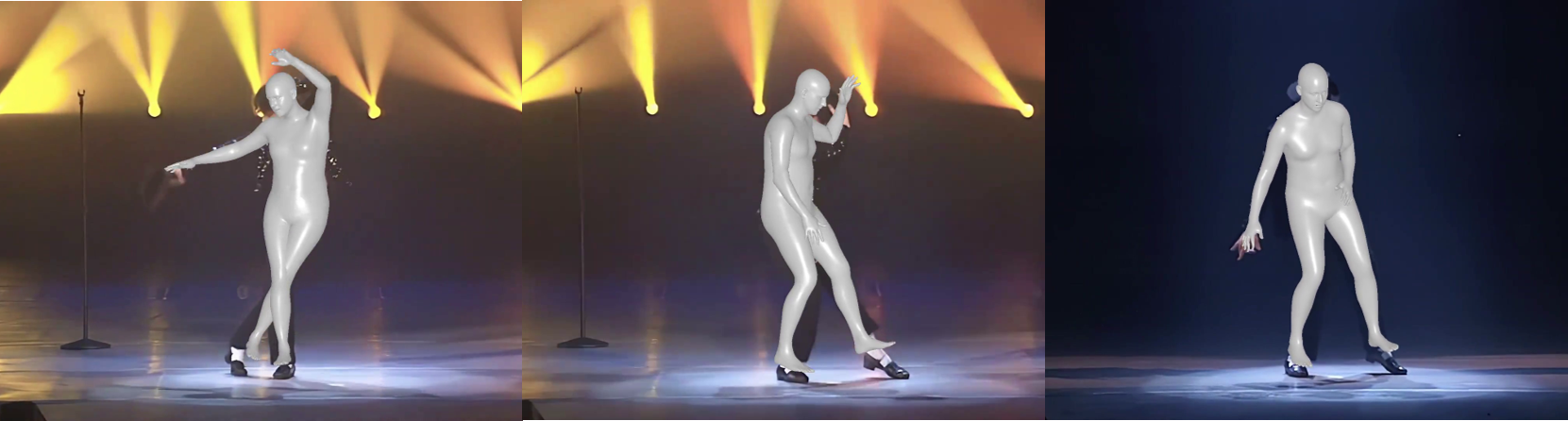}}
	\end{minipage} \\
    
    \hline
    
    \multirow{2}{*}{\thead{Motion Captioning \\ on Internet Videos \\ with GPT-4o~\cite{gpt-4o}}} & \begin{minipage}[b]{0.9\columnwidth}
		\centering
        {\includegraphics[width=1.\linewidth]{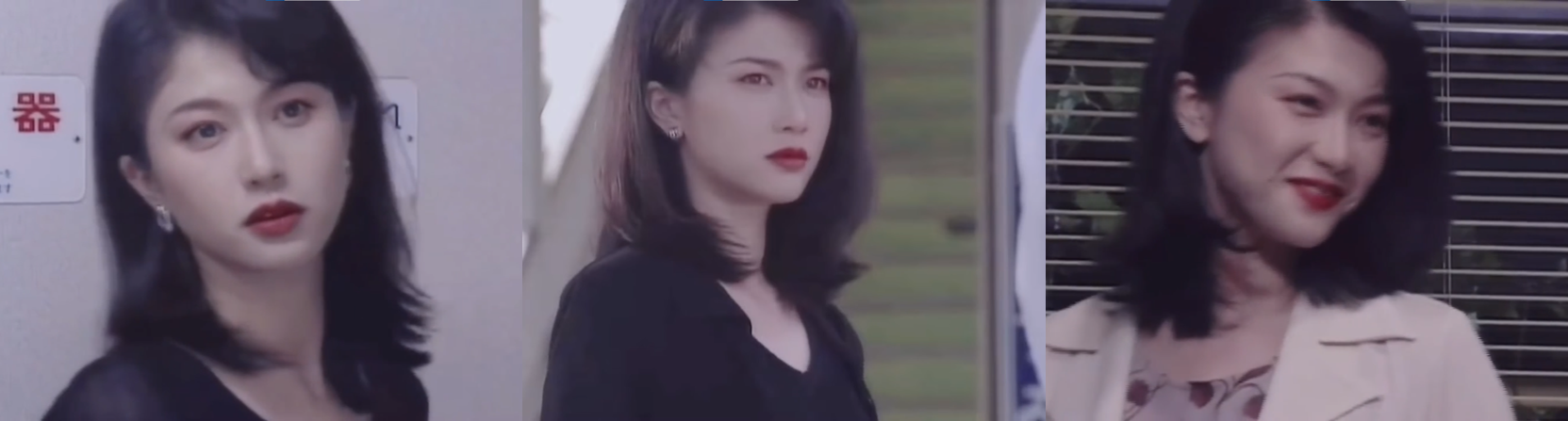}}
	\end{minipage} & \begin{minipage}[b]{0.9\columnwidth}
		\textit{1-3s: Turn head to the right and look straight ahead, with a neutral expression;
        4-5s: Turn body and look sideways, with a serious expression, almost no movement;
        6-8s: Turn to the left side, smiling while looking forward.}
	\end{minipage}\\
    
    & \begin{minipage}[b]{0.9\columnwidth}
		\centering
        {\includegraphics[width=1.\linewidth]{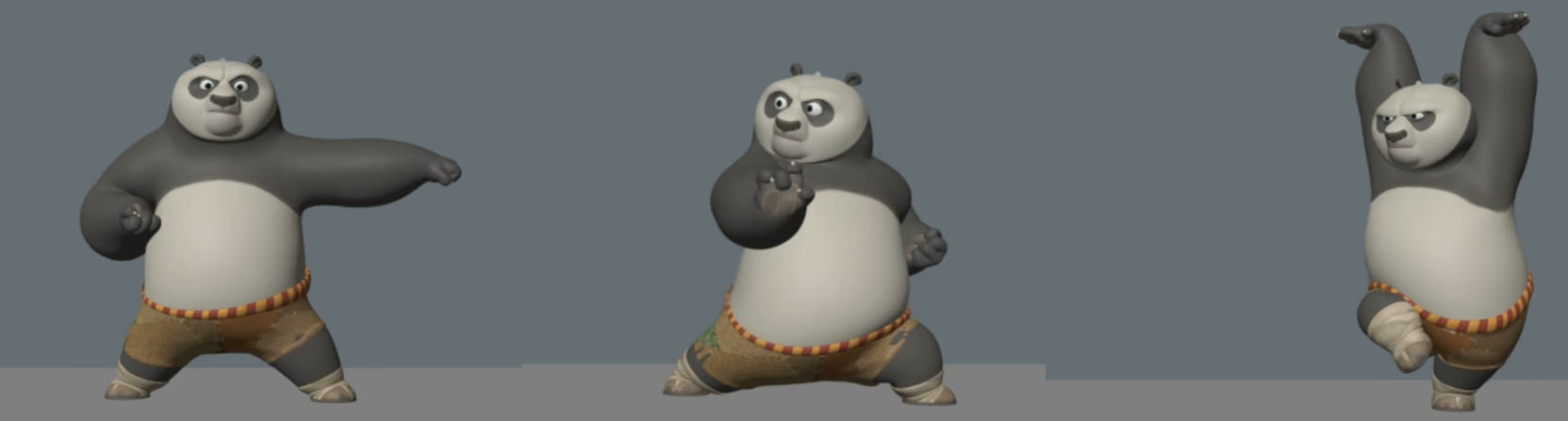}}
	\end{minipage}  & \begin{minipage}[b]{0.9\columnwidth}
		\textit{1-2s: A panda in a combat stance, right hand raised in a fist, left hand extended, with a serious facial expression; 3s: Panda's body tilts to the left side, right hand clenched in a fist, left hand stretched forward, eyes looking to the right front; 4-5s: Panda raises both hands above the head, lifting one leg.}
	\end{minipage} \\
    
    \hline
    
    \raisebox{.6\height}{\makecell[c]{\thead{Real Data \\ Collection from \\ VR Platforms}}} & \begin{minipage}[b]{0.9\columnwidth}
		\centering
        \raisebox{-.2\height}
        {\includegraphics[width=1.\linewidth]{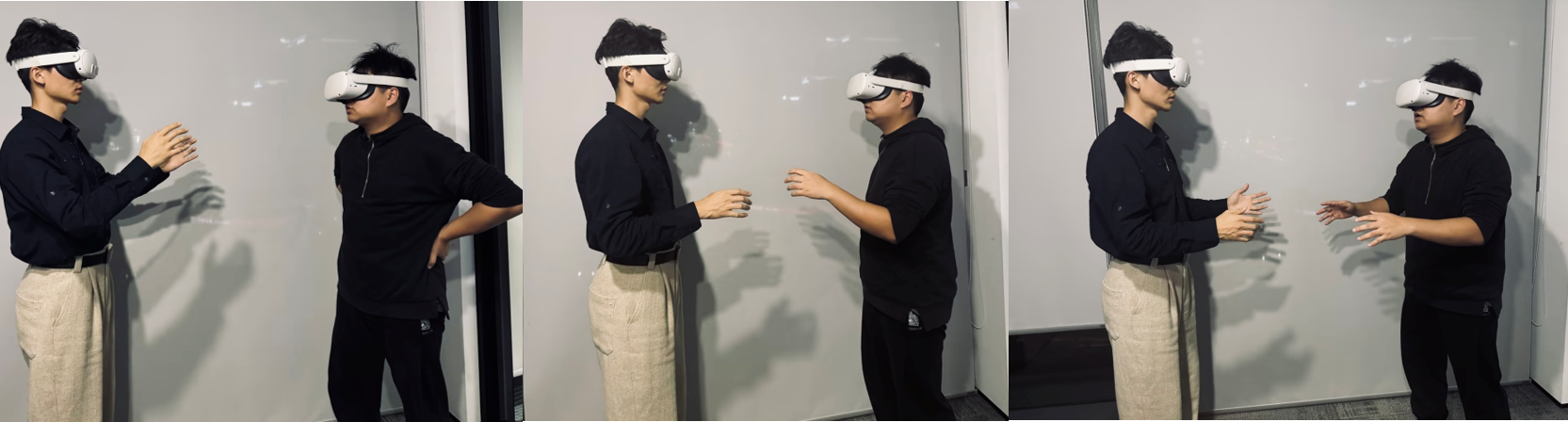}}
	\end{minipage} & \begin{minipage}[b]{0.9\columnwidth}
		\centering
        \raisebox{-.2\height}
        {\includegraphics[width=1.\linewidth]{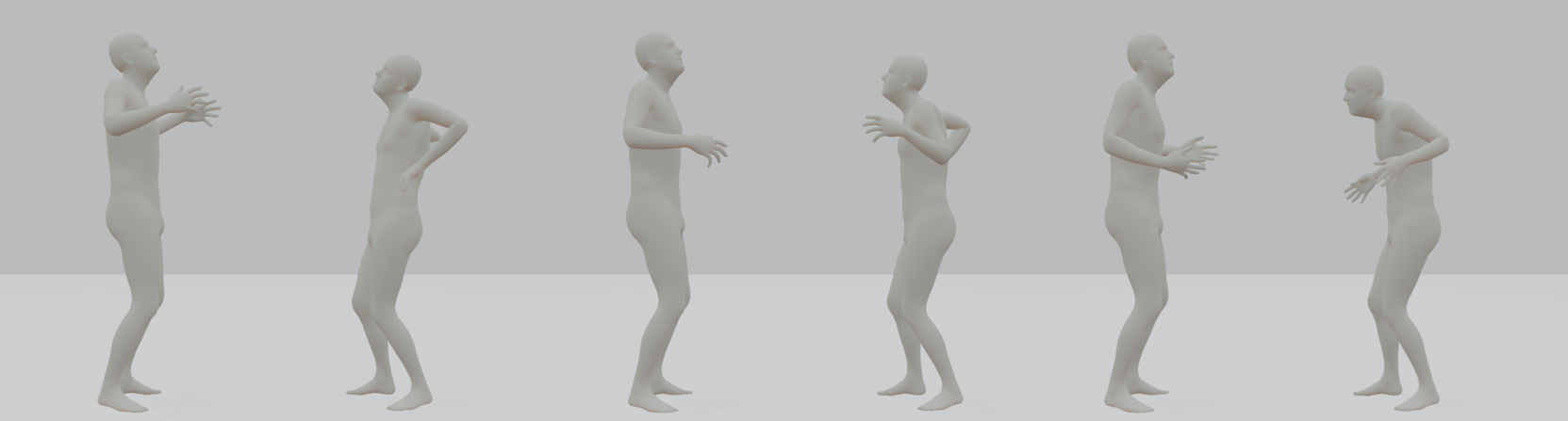}}
	\end{minipage} \\

    \hline

    \raisebox{.75\height}{\makecell[c]{\thead{Synthetic Data \\ Generation from \\ Existing Datasets}}} & \begin{minipage}[b]{0.9\columnwidth}
		\centering
        \raisebox{-.2\height}
        {\includegraphics[width=1.\linewidth]{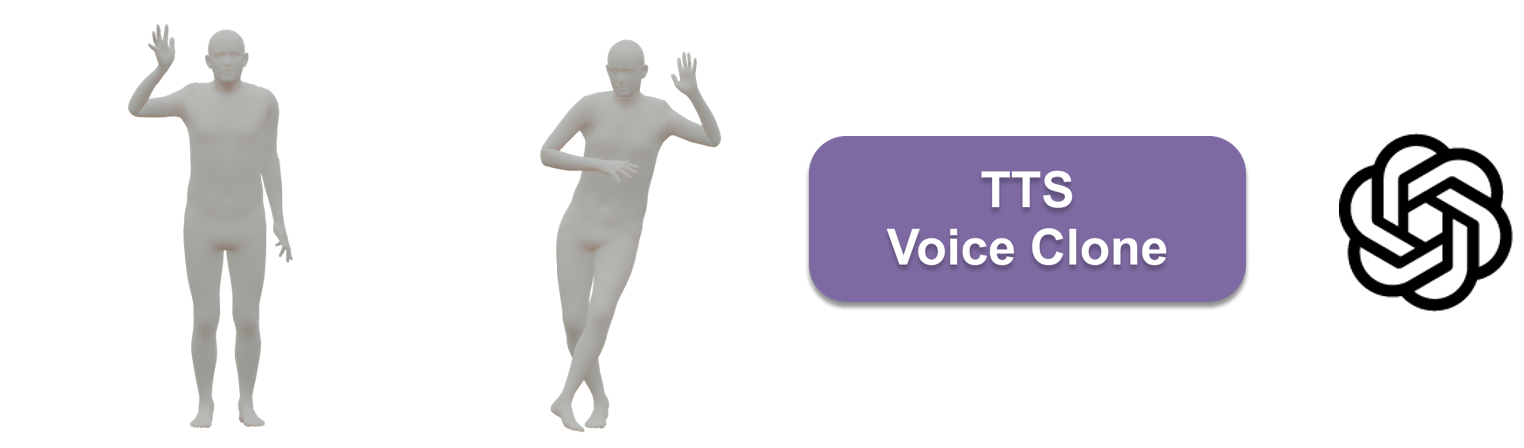}}
	\end{minipage} & \begin{minipage}[b]{1.\columnwidth}
		\centering
        \raisebox{-.2\height}
        {\includegraphics[width=.9\linewidth]{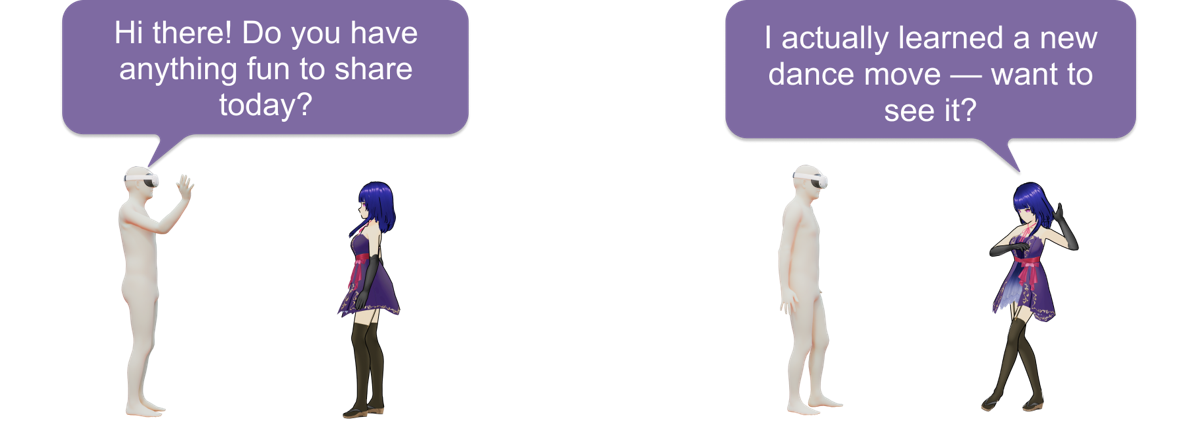}}
	\end{minipage} \\

    \bottomrule
    \end{tabular}}
    \label{table: data collection}
\end{table*}

%% file: figures/topics.tex
\begin{figure}[t]
    \centering
    \begin{subfigure}[t]{0.45\linewidth}
    \includegraphics[width=\linewidth]{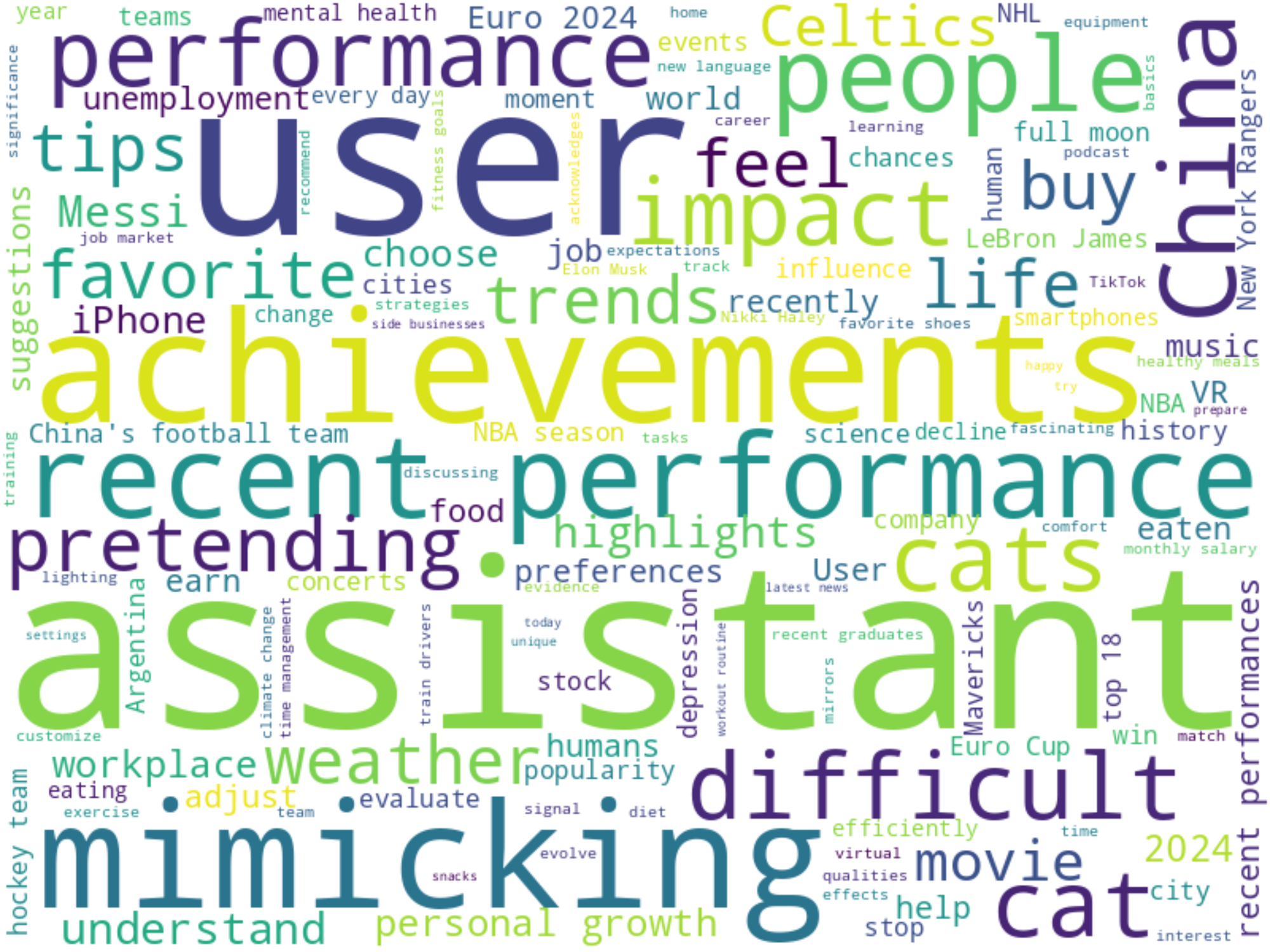}
    \caption{Samantha}
  \end{subfigure}
  \hspace{0.5cm}
   \begin{subfigure}[t]{0.45\linewidth}
    \includegraphics[width=\linewidth]{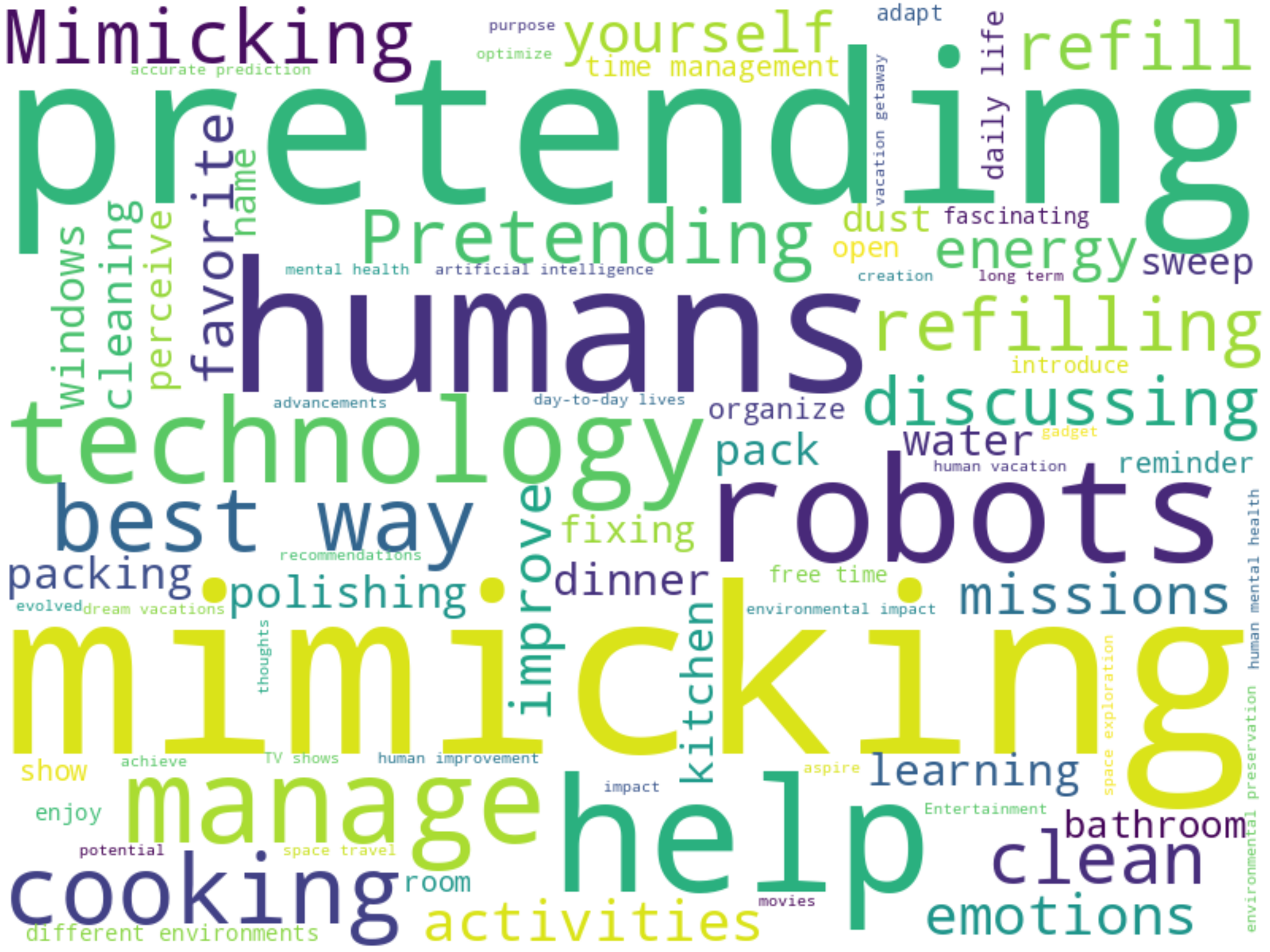}
    \caption{K-VRC}
  \end{subfigure}
  \hfill
      \centering
    \begin{subfigure}[t]{0.45\linewidth}
    \includegraphics[width=\linewidth]{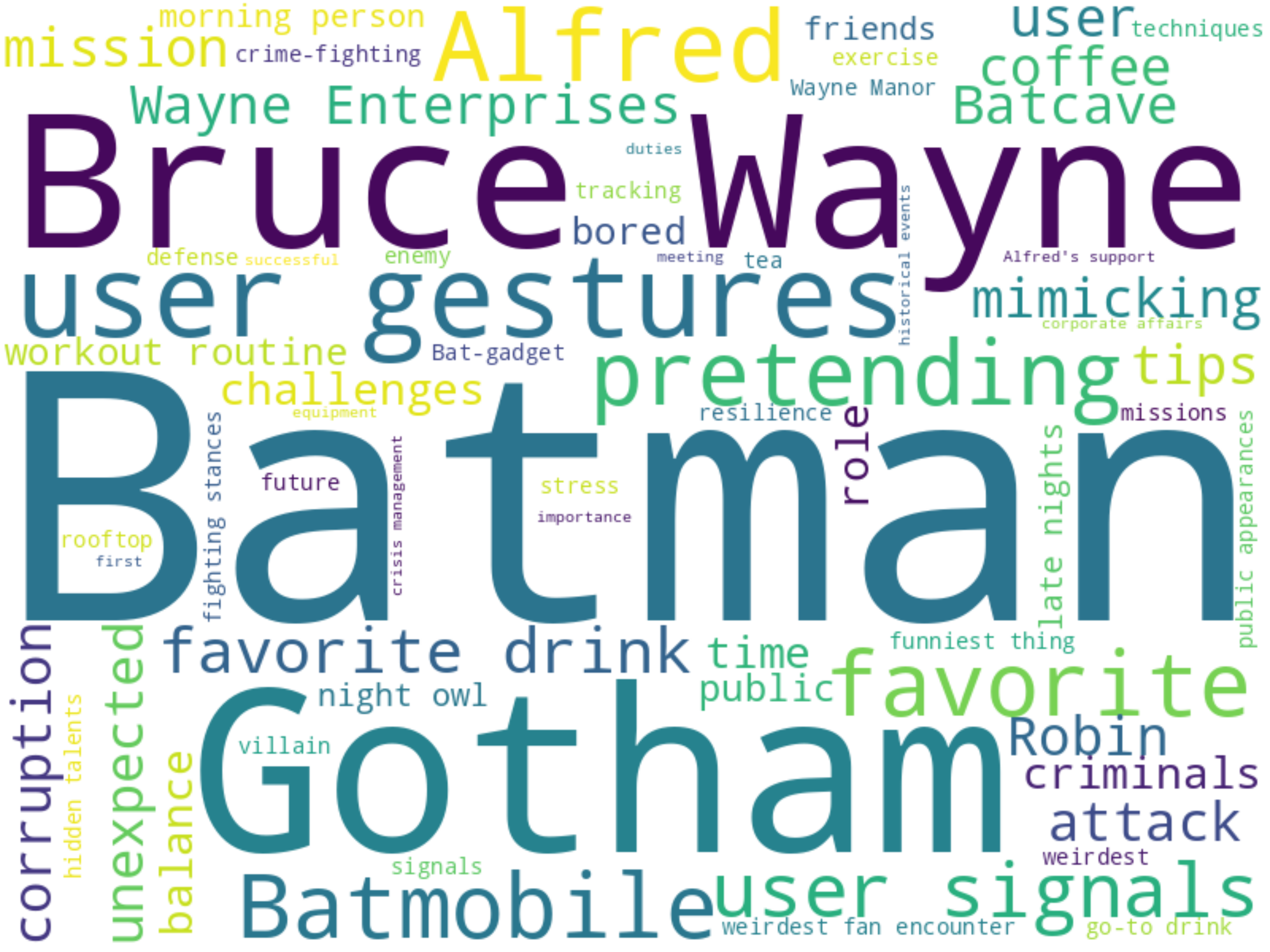}
    \caption{Batman}
  \end{subfigure}
  \hspace{0.5cm}
   \begin{subfigure}[t]{0.45\linewidth}
    \includegraphics[width=\linewidth]{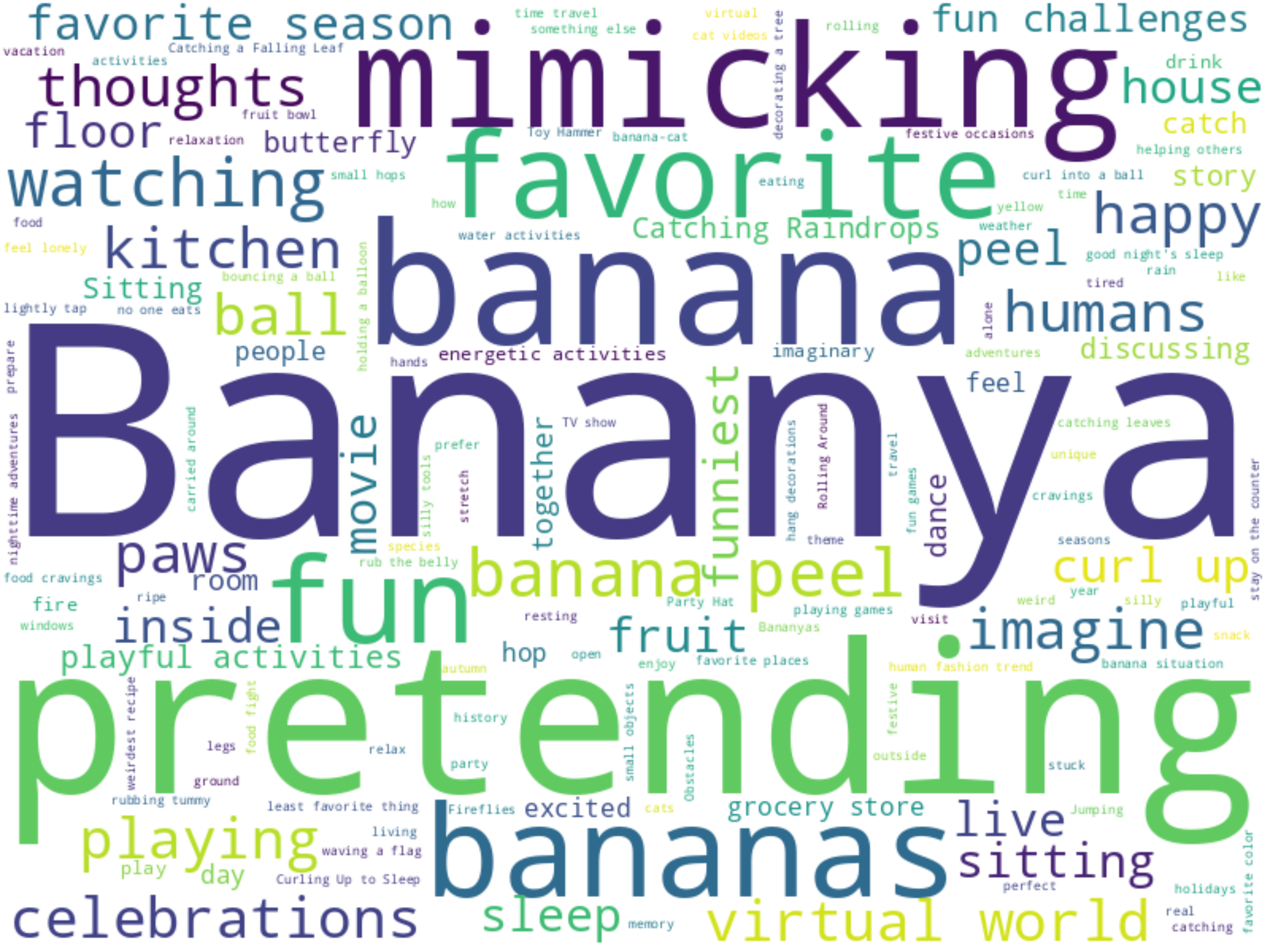}
    \caption{Banaya}
  \end{subfigure}
    \caption{Word cloud visualization of the keywords in the collected characters' topics.}
    \label{fig: supp: topics}
\end{figure}

%% file: sec/supp_D_experiments.tex
\section{More Details of Experiments}
\label{sec: supp: experimental settings}

\subsection{LLM Selection}
\label{sec: supp: llm selection}
We chose Llama2-7B~\cite{llama2} because at the time of our experiments, end-to-end models with speech pre-training were scarce, with AnyGPT~\cite{anygpt} being one of the few that performed well. 
Thus we selected the Llama2 series as the backbone for fair comparison in subsequent experiments. 
Readers aiming to achieve the best results can certainly choose state-of-the-art models as the backbone.

The Llama2-7B-chat model~\cite{llama2} tends to output increasingly longer dialogue content, which for \textit{LLM+Speech} methods results in high inference latency from both LLM and TTS (sometimes exceeding 30 seconds). 
Therefore, through post-processing, we truncate the output content to a maximum of 3 sentences. 
While truncating output content somewhat affects user experience, the lower user latency generally results in a better overall experience.

\subsection{Voice Cloning Comparison}
Voice cloning / TTS has numerous available products and open-source models in both industry and academia, each with different focuses. 
We aim to achieve the best voice cloning effect in near real-time conditions. 
For this purpose, we compare these software and algorithms: ElevenLabs Instant Voice Cloning~\cite{11labs}, ChatTTS + OpenVoice~\cite{chattts, openvoice}, XTTS\_v2~\cite{xtts}, MARS5~\cite{MARS5}, and Bark~\cite{bark}. 
Among them, MARS5~\cite{MARS5} uses a diffusion~\cite{ddpm} framework and is relatively slow; ElevenLabs~\cite{11labs} produces the best results but has high API costs and tends to generate speech at a faster pace. 
XTTS\_v2~\cite{xtts} is a more suitable option, and can achieve a good balance between speed and quality.

When \methodname{} processes speech, we use the pre-trained SpeechTokenizer~\cite{speechtokenizer} and SoundStorm~\cite{soundstorm} from AnyGPT~\cite{anygpt}. In SpeechTokenizer~\cite{speechtokenizer}, one second of speech is encoded into 400 tokens across 8 layers. We only select tokens from the first semantic layer (50 tokens in total) to send to \methodname{} for processing. 
During SoundStorm~\cite{soundstorm} decoding, we choose 4 to 6 seconds of voice prompt based on the character and generate the speech with 4 iteration steps.

%% file: sec/acknowledgements.tex
\section{Acknowledgments}
We extend our sincere gratitude to Fei Xia, Huazhe Xu, Tao Kong, Jiangyong Huang for their insights from the embodied intelligence field. 
We thank Mingyuan Zhang, Fangzhou Hong, and Xinying Guo for discussions on motion generation, Bo Li and Yuanhan Zhang for advice on multimodal model training.
We would also like to acknowledge Han Du, Fanzhou Wang, Jiaqi Li, Liang Pan, Peng Gao, and Yukun Wei for insightful discussions on the topic.